\documentclass[sigconf, balance=false]{acmart}

\usepackage{multirow}
\usepackage{graphicx}
\usepackage{array}
\usepackage{subcaption}
\usepackage{algorithm,algorithmic}
\usepackage{epstopdf}
\usepackage{parskip}
\usepackage{lipsum}
\usepackage{microtype}
\usepackage{booktabs} 
\setcopyright{none}
\renewcommand\footnotetextcopyrightpermission[1]{} 

\AtBeginDocument{%
  \providecommand\BibTeX{{%
    \normalfont B\kern-0.5em{\scshape i\kern-0.25em b}\kern-0.8em\TeX}}}

\begin{document}

\title{Free Lunch for Privacy Preserving Distributed Graph Learning}

\author{\textbf{Nimesh Agrawal}}
\email{eey217515@iitd.ac.in}
\vspace{-1em}
\affiliation{
  \institution{Department of Electrical Engineering}
  \institution{Indian Institute of Technology, Delhi}
  \state{Delhi}
  \country{India}
  \postcode{110016}
}
\author{\textbf{Nikita Malik}}
\email{bsz218185@dbst.iitd.ac.in}
\vspace{-1em}
\affiliation{
  \institution{Bharti School of Telecommunication Technology and Management}
  \institution{Indian Institute of Technology, Delhi}
  \state{Delhi}
  \country{India}
  \postcode{110016}
}
\author{\textbf{Sandeep Kumar}}
\email{ksandeep@iitd.ac.in}
\affiliation{%
  \institution{Department of Electrical Engineering}
    \institution{Yardi School of Artificial Intelligence}
    \institution{Bharti School of Telecommunication Technology and Management}
  \institution{Indian Institute of Technology, Delhi}
  \state{Delhi}
  \country{India}
  \postcode{110016}
}

\begin{abstract}
Learning on graphs is becoming prevalent in a wide range of applications including social networks, robotics, communication, medicine, etc. These datasets belonging to entities often contain critical private information. The utilization of data for graph learning applications is hampered by the growing privacy concerns from users on data sharing. Existing privacy-preserving methods pre-process the data to extract user-side features, and only these features are used for subsequent learning. Unfortunately, these methods are vulnerable to adversarial attacks to infer private attributes. We present a novel privacy-respecting framework for distributed graph learning and graph-based machine learning. In order to perform graph learning and other downstream tasks on the server side, this framework aims to learn features as well as distances without requiring actual features while preserving the original structural properties of the raw data. The proposed framework is quite generic and highly adaptable. We demonstrate the utility of the Euclidean space, but it can be applied with any existing method of distance approximation and graph learning for the relevant spaces. Through extensive experimentation on both synthetic and real datasets, we demonstrate the efficacy of the framework in terms of comparing the results obtained without data sharing to those obtained with data sharing as a benchmark. This is, to our knowledge, the first privacy-preserving distributed graph learning framework.

\end{abstract}



\maketitle
\pagestyle{plain}

\section{INTRODUCTION} \label{sec:Introduction}

Graphs are a powerful tool in data analysis because they can capture complicated structures that appear to be inherent in seemingly unstructured high-dimensional data. Graph data are ubiquitous in a plethora of real-world data including social networks, patient-similarity networks, and various other scientific applications. Graph Learning, in general, is emerging as an important problem \cite{kalofolias2016learn, dong2016learning, lake2010discovering} because of its potential applications in text, images, knowledge graphs, community detection \cite{liu2020graph}, recommendation systems \cite{fouss2007random}, fraud detection \cite{noble2003graph}. Most graph learning techniques are mainly designed for centralized data storage. However, for many areas, the data is spatially and geographically distributed. For example, user data in social networks, clinical features of patients in Patient Similarity Networks (PSN), data in transportation systems, etc. Collecting data from devices scattered across the globe becomes challenging and often arises privacy risks as the data contains private information. The data collector may misuse the data, or it may be exposed to adversarial attacks. Data security and privacy is a major hurdle for many big companies. After the major data breach of Facebook \cite{cadwalladr2018revealed}, for example, many users considered sharing their personal attributes as an intrusion into their privacy. These raising privacy concerns hinder the use of these isolated data for graph analytics-based applications.\par

There are some privacy critical applications where it is illegal to share sensitive data among different entities or organizations. Genomic data, for example, have been shown to be sensitive, necessitating additional safety measures before sharing for machine learning procedures \cite{AZIZ2022104113}. Patient information is found in a variety of forms and sources (hospitals,
home-based devices, and patients' mobile devices). PSN can use a variety of measurements, such as laboratory test results and vital signs, to create a multidimensional representation of each patient \cite{el2021federated}. These datasets are complicated, thus learning graph topologies while protecting privacy is necessary in order to get insights from such a large amount of data. The increasing growth in privacy awareness has made research into distributed machine learning more pertinent. Google's GBoard makes use of Federated Learning (FL) for next-word prediction \cite{hard2018federated}, Apple is using FL for applications like the vocal classifier for “Hey Siri” \cite{siri}, Snips has explored FL for hotword detection \cite{leroy2019federated}.\par

Existing privacy-protection technologies struggle to strike a balance between privacy and utility. A common approach is to convert the original data into features that are specific to a certain task and then only send these task-oriented features to a cloud server, instead of the raw data. Recent advancements in model inversion attacks \cite{dosovitskiy2016inverting, mahendran2015understanding} have shown that adversaries can reverse engineer acquired features back to raw data, and use it to infer sensitive information. FL is also a popular framework for distributed machine learning which aggregates model updates from clients' parameters. But according to Zhu and Han's study \cite{zhu2019deep}, it is found that model gradients can be used in some scenarios to partially reconstruct client data. 
The requirement for graph learning in a distributed environment that protects user privacy has motivated us for the development of a privacy-respecting distributed graph learning framework.\par

\begin{figure*}[h]
    \centering
    \includegraphics[width=2.1\columnwidth]{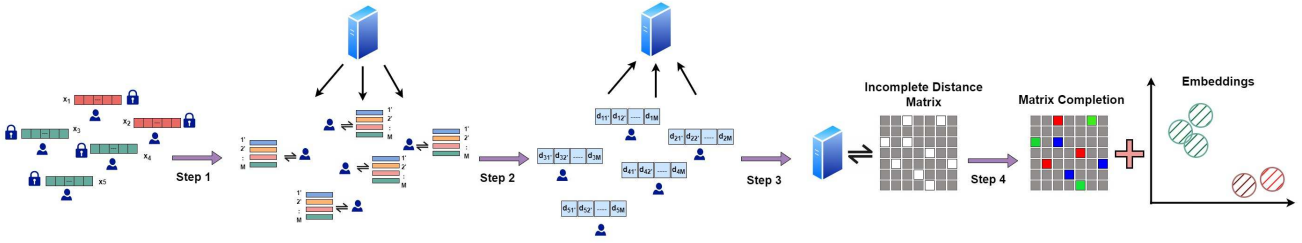}
    \caption{The overall framework of PPDA. Each user in the distributed network has their attributes privately stored with them. In Step ($\textbf{1}$), a centralized server populates all the users with a predefined set of feature vectors called `\textbf{Anchors}'. Anchor features have the same distribution as the underlying user data. In Step ($\textbf{2}$), all the users compute the pairwise distance with all the anchors and push it to the server. To this end, at the server, we have an Incomplete Distance Matrix with all the user-anchor distances known and user-user distances unknown as seen in Step ($\textbf{3}$). Finally, in Step ($\textbf{4}$), we obtain the completed distance matrix along with embeddings which are used further for graph learning and downstream tasks.}
    \label{fig:framework}
\end{figure*}

In this paper, we present a novel privacy-preserving distributed graph learning framework, PPDA for graph learning that respects user privacy. The ultimate objective of this framework is to enable graph-based learning and analysis of distributed data without requiring users to share their data with a central server. Any graph, represented by an Adjacency matrix ($A$), needs the pairwise distance between nodes or the sample covariance matrix $S$ for learning $A$ \cite{kalofolias2016learn, kalofolias2017large,egilmez2017graph, kumar2020unified}. It becomes challenging to compute inter-user distance keeping privacy in consideration. To resolve it, as shown in Figure \ref{fig:framework}, we first populate all the users/clients with a set of known attributes, which we refer in this paper as 'Anchors'. Each client sends their distance with all the anchors back to the server after local gossiping with anchors. As user-to-user distance is still unknown, we have an incomplete distance matrix on the server. We then perform matrix completion to estimate inter-user distance and also learn a feature representation of users such that the underlying neighborhood structure in the raw data is maximally retained. We consider only the Euclidean space for demonstration in this paper, but the framework can be utilized for any embedding space such as Riemannian, spherical or hyperbolic space \cite{wilson2014spherical, lindman1978constant, elad2005texture}. Extensive experimentation on synthetic and real datasets elucidates that our method can be built on top of all the existing pipelines for matrix completion and graph learning frameworks.\par

\textbf{Notations:} The lower case (bold) letters denote scalars (vectors) and upper case letters denote matrices. The $(i, j)$ - th entry of matrix $X$ is denoted by $X_{ij}$. The pseudo-inverse and transpose of matrix $X$ are denoted by $X^{\dagger}$ and $X^{T}$ respectively. The all-one vectors of appropriate size are denoted by $\textbf{1}$. The Frobenius norm of a matrix $X$
is defined by $\left \| X \right \|_{F}^{2} = \sum_{ij}^{}X_{ij}^2$. The Euclidean norm of vector $\boldsymbol{x}$ is denoted as $\left \| \boldsymbol{x} \right \|_2$.\par

The rest of the paper is organized as follows. In section \hyperref[sec: Related work]{2}, we briefly review the related work. Section \hyperref[sec:background]{3} describes the problem statement and related background. Section \hyperref[sec: PPDA framework]{4} provides a detailed explanation of the PPDA framework, including the underlying theory and the parameters used in our method. PPDA for graph-based clustering is presented in Section \hyperref[sec:clustering]{5}. Section \hyperref[sec:experiments]{6} showcases the effectiveness of our framework through extensive experimentation, using various measures and evaluates the performance of the method for downstream applications. In the final Section \hyperref[sec:conclusion]{7}, the conclusion is discussed.

\section{RELATED WORK} \label{sec: Related work}
Several techniques of Privacy-Preserving Data Mining (PPDM) have been developed to extract knowledge from large datasets while keeping sensitive information confidential \cite{aggarwal2008general}. Most of these techniques are based on perturbing original data including k-anonymity \cite{sweeney2002k}, t-closeness\cite{li2006t}, and l-diversity \cite{machanavajjhala2007diversity}. The aforementioned methods are intended for safeguarding sensitive attributes in centralized data storage and therefore, are not appropriate for addressing our problem. Differential privacy (DP) \cite{smith2017interaction, wang2017locally, huang2020dp}and Homomorphic encryption (HE) \cite{rivest1978data} are other types of widely adopted methods that ensure strong privacy. DP aims to achieve privacy by adding noise to actual data samples in such a way that the likelihood of two different true data samples generating the same noisy data sample is similar. On the other hand, HE allows mathematical operations to be performed on encrypted data, resulting in an encrypted outcome that, when decrypted, is equivalent to the outcome of performing the same operations on unencrypted data. However, both of these techniques come with their own drawbacks. DP can lead to utility loss and HE can require a large amount of computational resources, making it impractical in certain scenarios. \par

As it is necessary to compare signals to get the similarity graph, various methods for privacy-preserving distance computation are presented in \cite{rane2013privacy}. A new approach for highly efficient secure computation for computing an approximation of the Similar Patient Query problem is proposed in \cite{asharov2017privacy}. However, these methods do not take into account the distributed data. A recent study \cite{nobre2022distributed}, presents a new method for distributed graph learning that takes into account the cost of communication for data exchange between nodes in the graph. The downside of this algorithm is that it does not consider data privacy. The bottlenecks of these methods motivate us to propose a framework that is simple enough and provides more general privacy protection.

\section{Background and Problem Statement} \label{sec:background}
In this section, we give a brief background of graphs and graph learning including probabilistic graphical models. Then we describe the problem statement and the parameters of our method.  

\subsection{Graph} \label{sec:graph} A graph with features is represented by $\mathcal{G} = (V,E,A,X)$ where $V = \left\{v_1, v_2,\cdots, v_N\right\}$ is the set of $N$ vertices, $E$ is the set of edges which is a subset of $V \times V$ and $A$ is the adjacency (weight) matrix. We take into account a basic undirected graph without a self-loop. The entries of weight matrix are defined as: $A_{ij} > 0$, if $(i, j) \in E$ and $A_{ij} = 0$, if $(i, j) \notin E$. We also have a feature matrix
$X \in \mathbb{R}^{N \times d} =  \left [ \boldsymbol{x_1}, \boldsymbol{x_2},\cdots,\boldsymbol{x_N} \right ]^T$ , where each row vector $\boldsymbol{x_i} \in \mathbb{R}^d$ is the feature vector associated with one of $N$ nodes of the graph $\mathcal{G}$. Consequently, each of the $d$ columns of $X$ can be viewed as a signal on the same graph. The Laplacian ($\Theta$) and Adjacency ($A$) matrices are often used to represent graphs, whose entries represent the connections between the vertices (nodes) of the graph. A diagonal matrix $Q$ with entries $Q_{ii} = \sum _{j}{A_{ij}} $ represents the degree of nodes of $\mathcal{G}$.

\subsection{Graph Learning} \label{sec:graph learning}
A basic premise about data residing on the graph is that it changes smoothly between connected nodes. To quantify this assumption, if two signals $\boldsymbol{x_{i}}$ and $\boldsymbol{x_{j}}$ are from the smooth set and are located on well-connected nodes (i.e $A_{ij}$ is large), then it is expected that their distance $\left \| \boldsymbol{x_i} - \boldsymbol{x_j}\right \|$ will be small, which leads to a small value for $\operatorname{tr}(X^T\Theta X)$, where $\Theta$ is the laplacian matrix defined as $Q - A$ \cite{kipf2016semi}. The matrix $\Theta \in \mathbb{R}^{N \times N}$ is a valid graph Laplacian matrix if it belongs to the following
set:
\begin{equation}
\mathcal{S}_{\Theta}=\left\{\Theta \in \mathbb{R}^{N \times N} \mid \Theta_{i j}=\Theta_{j i} \leq 0 \text { for } i \neq j ; \Theta_{i i}=-\sum_{j \neq i} \Theta_{i j}\right\}
\label{eq:14}
\end{equation}

The graph Laplacian matrix is a widely used tool in graph analysis and has many important applications in various fields. Some examples of its use include embedding, manifold learning, spectral sparsification, clustering, and semi-supervised learning \cite{zhu2003semi,belkin2006manifold,von2007tutorial}. Under this smoothness assumption, to learn a graph from data, \cite{kalofolias2016learn} proposed an optimization formulation in terms of adjacency matrix ($A$) and pairwise distance matrix ($K$), where $K_{ij} = \left \| \boldsymbol{x_i} - \boldsymbol{x_j}\right \|^2 
= d_{ij}$ represents the Euclidean distance between samples $\boldsymbol{x_{i}}$ and $\boldsymbol{x_{j}}$:

\begin{equation}
    \min_{A \, \in \, \mathcal{S}_{A}} \sum_{i = 1}^{N} A_{ij}d_{ij} +f(A)
    \label{eq:15}
\end{equation}
\begin{equation}
  \mathcal{S}_{A} =   \left \{ A \in \mathbb{R}_{+}^{N \times N}: A = A^T, \text{diag}(A) = 0 \right \}
  \label{eq:16}
\end{equation}

where $f(A)$ is the regularization term for limiting $A$ to satisfy certain properties, and $\mathcal{S}_{A}$ in eq. \eqref{eq:16} refers to the set for valid Adjacency matrix. 

\subsection{Probabilistic Graphical Models} \label{sec:PGM}

Conditional dependency relationships between a set of variables are encoded via Gaussian graphical modeling (GGM). In an undirected graphical topology, a GGM simulates the Gaussian property of multivariate data \cite{https://doi.org/10.48550/arxiv.1111.6925}. 
Let $\boldsymbol{x} =(x_1,...,x_N)^T$ be a $N$-dimensional zero mean multivariate random variable
associated with an undirected graph and $S \in R^{N \times N}$  be the sample covariance matrix (SCM) calculated from $d$ number of observations. 

In a GGM, the existence of an edge between two nodes
indicates that these two nodes are not conditionally independent given other
nodes. The GGM method learns a graph via the following optimization problem \cite{friedman2008sparse}:
\begin{equation}
    \max_{\Theta \, \in \, S^N_{++}} \log det(\Theta) - tr(XX^T \Theta) - \alpha h(\Theta)
\label{eq:22}   
\end{equation}
where $X \in \mathbb{R}^{N \times d}$ denotes the feature matrix, $\Theta \in R^{N \times N}$ denotes the graph matrix to be estimated, $N$ is the number of nodes (vertices) in the graph, $S^N_{++}\in R^{N \times N} $ denotes the set of positive definite matrices, $h(.)$ is a regularization term, and $\alpha >0$ is the regularization hyperparameter. The SCM given by $S = XX^T$ can also be written in terms of distance matrix $K$ given by:
\begin{equation}
S = \frac{-1}{2}\left ( K - \textbf{1}(\textbf{diag}(XX^T))^T - \textbf{diag}(XX^T)\textbf{1}^T \right )
\label{eq:26}   
\end{equation}

where \textbf{1} denotes the column vector of all ones and $\textbf{diag}(A)$ is a column vector of the diagonal entries of $A$.  Now, if the observed data is distributed according to a zero-mean $N$-variate Gaussian distribution, then the optimization in \eqref{eq:22} corresponds to the penalized maximum likelihood estimation (MLE) of the inverse covariance (precision) matrix of a Gaussian random vector also known as Gaussian Markov Random Field (GMRF) \cite{JMLR:v21:19-276}.
Let $\Theta$ be a $N \times N$ symmetric positive semidefinite matrix with rank $\left(N-k\right) >0$.
Then $ \boldsymbol{x} =(x_1,...,x_N)^T$ is an improper GMRF (IGMRF) of rank $\left(N-k\right)$.\par

Now, it is strongly evident from equations \eqref{eq:15} and \eqref{eq:22} that in order to learn a graph structure, we need to have the pairwise distances $d_{ij} = f(\boldsymbol{x_i}, \boldsymbol{x_j})$ to estimate $A$ or $\Theta$. However, access to the features is restricted due to privacy concerns. As a result, our goal is to find an approximation of the distances without exposing the features, which can then be used to compute the SCM using equation \eqref{eq:26}. The estimated distances and corresponding $S$ are then used further for graph learning and other related tasks while ensuring privacy protection.

\section{Proposed Privacy Preserving Distance Approximation (PPDA)} \label{sec: PPDA framework}
In this section, we formalize our problem setting, and then we introduce the proposed framework. Finally, we provide an illustration of the framework using Multi-dimensional Scaling (MDS) for embedding.

\subsection{Problem Formulation}\label{sec:problem_formulation}

Consider a distributed setting of $N$ isolated clients. Let $X_{NA} = \left [ \boldsymbol{x_1}, \boldsymbol{x_2},.....,\boldsymbol{x_N} \right ]^T \in \mathbb{R}^{N\times d}$ be the feature representation of these clients referred to as \textit{Non-anchors (NA)}, where $\boldsymbol{x_{i}} \in \mathbb{R}^{d}$ is the feature vector of dimension $d$ for $i^{th}$ user. A centralized server now generates $M$ reference nodes represented by $\left\{a_1, a_2, \cdots, a_M\right \}$ referred here as \textit{Anchors (A)} and sends the features associated with these anchors to all the clients $i \in \left \{1, 2,\cdots, N\right \}$. For instance, this server could be owned by companies such as Google or Facebook, healthcare institutions, or a government agency. This is shown as Step ($\textbf{1}$) in Figure \ref{fig:framework}. Let  $X_{A} = \left [ \boldsymbol{p_1},\boldsymbol{p_2},\cdots,\boldsymbol{p_M} \right ]^T \in \mathbb{R}^{M\times d}$ be the feature representation of these anchors. Now each client $i$ have access to its own feature and anchor features i.e $\left\{\boldsymbol{x_i}, \boldsymbol{p_1}, \boldsymbol{p_2}, \cdots, \boldsymbol{p_M}\right \} $. Each client then computes the exact distance with each anchor given by $\left(d_{ia_1}, d_{ia_2}, \cdots, d_{ia_M}\right ) \forall i \in \left\{1, 2, \cdots, N\right\}$ and sends these distances back to server shown as Step ($\textbf{2}$) in Figure \ref{fig:framework}. As a result, we have an incomplete distance matrix, with all client-to-client distances unknown. This matrix is then used to perform matrix completion to obtain an inter-client distance matrix for learning a graph and conducting graph-based downstream tasks.

\subsubsection{Construction of Incomplete Distance matrix for PPDA} \label{sec:construction}

As mentioned in section \hyperref[sec:problem_formulation]{4.1}, using the pairwise distance between clients and anchors, we construct a distance matrix $D$ given in equation \eqref{eq:8}, which can be partitioned into three block matrices namely: $K$ being the unknown client-to-client distance, $D_{12}$ being the client to anchor distance and $D_{22}$ being the anchor to anchor distance. 
\begin{equation}
   D =  \begin{bmatrix}
K & D_{12}\\ 
D_{12}^T & D_{22}
\end{bmatrix}
\label{eq:8}
\end{equation}

\subsection{Anchor Generation} \label{sec:anchor_schedule}
A crucial aspect of the framework to take into account is the proper scheduling of the anchors. It has a significant impact on the overall performance and accuracy of the framework. There are several factors to consider when developing the anchor scheduling strategy, including:\par

\textbf{Number of anchors}: The number of anchors used in the framework has a direct impact on the algorithmic performance. Too few anchors may not preserve the structural information while ensuring privacy, while too many anchors may lead to overfitting and may violate privacy. \par

\textbf{Selection criteria}: The criteria used to select anchors can also impact the performance of the system. Selecting anchors from the same probability distribution as of the underlying user data may be more effective than selecting them at random. For example, the data distribution of patient similarity networks or social networks will depend on factors including a number of patients/users or similarity of patients/connection between users. \par

The pseudo-code for PPDA is listed down in Algorithm \ref{alg:ppda_algo}. It is evident that the proposed framework is straightforward as it only requires a set of anchors to approximate the inter-client distance, thereby providing privacy protection at no extra cost.
\begin{algorithm}
 \caption{PPDA: Privacy Preserved Distance Approximation}
 \begin{algorithmic}[1]
 \renewcommand{\algorithmicrequire}{\textbf{Input:}}
 \renewcommand{\algorithmicensure}{\textbf{Output:}}
 \REQUIRE M anchors $X_{A} \in \mathbb{R}^{M\times d}$, embedding dimension ($\widetilde{d}$)
 \ENSURE Embedding $\left(\widetilde{X}\right)$ in dimension $\widetilde{d}$, Estimated user to user distance matrix $K \in \mathbb{R}^{N \times N}$
  \STATE Populate all the users with anchors $X_{A}$ from the server.
  \STATE Compute anchor-anchor distance matrix $D_{22}$
  \FOR{every user $i \in \left \{1,2,\cdots,N \right \}$} \STATE Compute user to $M$ anchors distance to get matrix $D_{12}$ \ENDFOR
  \STATE Send the incomplete distance matrix formed as in \eqref{eq:8} back to the server.
  \STATE Perform distance matrix completion to obtain $K$ along with $\widetilde{X} \in  \mathbb{R}^{N\times \widetilde{d}}$

 \end{algorithmic}
 \label{alg:ppda_algo}
 \end{algorithm}
 
\subsection{Demonstrating PPDA using MDS} \label{sec:ppda using mds}

Utilizing the measurements of distances among pairs of objects, MDS (multidimensional scaling) finds a representation of each object in $d$ - dimensional space such that the distances are preserved in the estimated configuration as closely as possible.
To validate the goodness-of-fit measure, MDS optimizes the loss function (known as "Stress"($\sigma$)) given by:

\begin{equation}
    \sigma(X) = \min_{X}\sum_{i<j\leq N} w_{ij} \left ( \delta_{ij} - d_{ij}(X) \right )^{2}
    \label{eq:1}
\end{equation}

\begin{equation}
w_{i j}=\left\{\begin{array}{l}
1, \text { if } \delta_{i j} \text { is known } \\
0, \text { if } \delta_{i j} \text { is missing }
\end{array}\right.
\label{eq:2}
\end{equation}

\begin{equation}
  W =   \left[\begin{array}{cc}
{0}_{N, N} & 11_{N, M} \\
11_{N, M}^T & 11_{M, M}
\end{array}\right]
\label{eq:3}
\end{equation}

where $X$ represents the computed configuration, $d_{ij}(X) = \left \| \boldsymbol{x_{i}} - \boldsymbol{x_{j}} \right \|$ is the Euclidean distance between nodes $i$ and $j$, $\delta_{ij}$ is the measured distance  computed privately and weights $w_{ij}$ are defined in equation \eqref{eq:2}. Placing the weights of unknown inter-user distance to zero, the weight matrix $W$ can be partitioned into block matrices as shown in equation \eqref{eq:3}, where $11_{N,M}$ is a matrix of ones with shape $N\times M$. De Leeuw \cite{de2005applications} applied an iterative method called SMACOF (Scaling by Majorizing a Convex Function) to estimate the configuration $X$. As the objective is a non-convex function, SMACOF minimizes the stress using the simple quadratic function $\tau(X, Z)$ which bounds $\sigma(X)$ (the complicated function) from above and
meets the surface at the so-called supporting point $Z$ as given in equation \eqref{eq:4}.

\begin{equation}
\begin{aligned}
\sigma(X) \leq & \tau(X, Z)=\sum_{i<j} w_{i j} \delta_{i j}^2+\sum_{i<j} w_{i j} d_{i j}^2(X) \\
& -2 \sum_{i<j} w_{i j} \delta_{i j}{ }^2 \frac{\left(\boldsymbol{{x}_i}-\boldsymbol{{x}_j}\right)^T\left(\boldsymbol{{z}_i}-\boldsymbol{{z}_j}\right)}{\left\|\boldsymbol{{z}_i}-\boldsymbol{{z}_j}\right\|}
\end{aligned}
\label{eq:4}
\end{equation}

Equation \eqref{eq:4} can be written in matrix form as:
\begin{equation}
\tau({X}, {Z})={C}+\operatorname{tr}\left({X}^T {V X}\right)-2 \operatorname{tr}\left({X}^T {B}({Z}){Z}\right)
\label{eq:5}
\end{equation}

The iterative solution which guarantees monotone convergence of stress \cite{de1988convergence} is given by equation \eqref{eq:6}, where ${Z} ={X}^{k-1}$:
\begin{equation}
{X}^{(k)}=\min _X \tau({X}, {Z})={V}^{\dagger} {B}({X}^{(k-1)}) {X}^{(k-1)}
\label{eq:6}
\end{equation}
This algorithm offers flexibility to embed features in any dimension other than $d$, which enables the handling of high-dimensional data and also meets privacy constraints. As $V$ is not of full rank, hence the Moore-Penrose pseudoinverse $V^{\dagger}$ is used. The elements of the matrix $B(X)$ and $V$ are defined in equation \eqref{eq:7}. 
\begin{equation}
\begin{gathered}
b_{i j}=\left\{\begin{array}{lr}
-\frac{w_{i j} \delta_{i j}}{d_{i j}({X})} & d_{i j}({X}) \neq 0, i \neq j \\
0 & d_{i j}({X})=0, i \neq j \\
-\sum_{j=1, j \neq i}^N b_{i j} & i=j
\end{array}\right. \\
v_{i j}= \begin{cases}-w_{i j} & i \neq j \\
-\sum_{j=1, j \neq i}^N v_{i j} & i=j\end{cases}
\end{gathered}
\label{eq:7}
\end{equation}

Next, as the features of anchors are being shared among all users, it is shown that only user-level embedding can be computed as a function of $X_{A}$ \cite{di2017multidimensional}, utilizing the incomplete distance matrix given in equation \eqref{eq:8}.
\begin{figure*}[h]
    \centering
    \includegraphics[width=2.1\columnwidth]{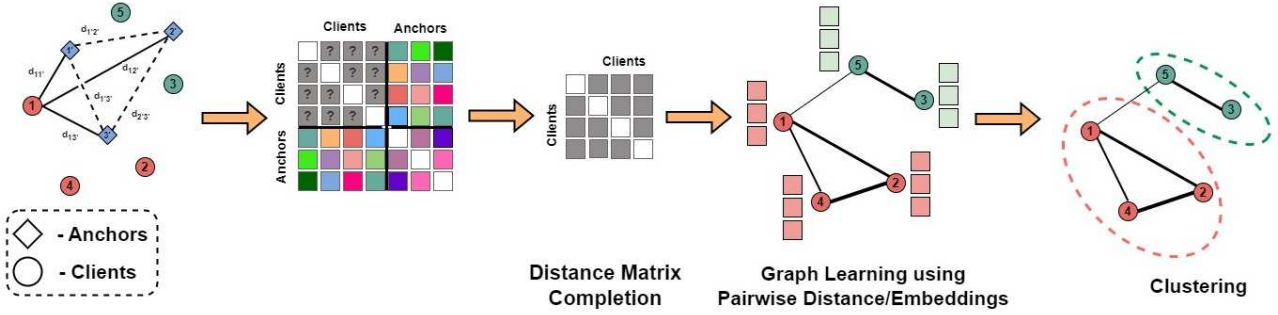}
    \vspace{-1em}
    \caption{The overall pipeline for Private Graph-based Clustering. For the client's and anchor's attributes lying in any feature space, we first compute all the pairwise distances such that only inter-client distance is unknown. Next, we form an incomplete distance matrix in block form as shown in the figure. Using any existing methods for distance matrix completion we estimate unknown distances as well as the client's feature representation. Based on this pairwise distance/embedding, a graph structure among the clients is learnt, which is further used for graph-based clustering or any other downstream task.}
    \label{fig:pipeline}
\end{figure*}
\subsection{Using Anchored-MDS} \label{sec:anchor mds}

For a network of $N$ users with $M$ anchors, with total $(N + M)$ unique nodes, we can partition $X$ from equation \eqref{eq:6} as:
\begin{equation}
\begin{aligned}
{X} & =\left[\begin{array}{l}
{X}_{NA} \\
{X}_A
\end{array}\right], \quad \text { with } \\
{X}_{NA} & =\left[\boldsymbol{x}_1, \cdots, \boldsymbol{x}_N\right]^T \in \mathbb{R}^{N \times d} \\ 
{X}_A & =\left[\boldsymbol{x}_{N+1}, \cdots,\boldsymbol{x}_{N+M}\right]^T \in \mathbb{R}^{M \times d}
\end{aligned}
\label{eq:9}
\end{equation}
Similarly, matrices $V$ and $B(X)$ as described in \eqref{eq:7} can also be partitioned into block matrices as shown in equation \eqref{eq:10}. It should be noted that $V$ is the Laplacian of the weight matrix $W$ defined in \eqref{eq:3}, whereas $B(X)$ is the Laplacian of weighted $W$. From this definition, $V$ can be represented in the block form as given in \eqref{eq:11}.
\begin{equation}
    V = \begin{bmatrix}
{V_{11}} & {V_{12}}\\ 
{V_{12}^T} & {V_{22}}
\end{bmatrix}, \, B(X) = \begin{bmatrix}
{B_{11}} & {B_{12}}\\ 
{B_{12}^T} & {B_{22}}
\end{bmatrix}
\label{eq:10}
\end{equation}

\begin{equation}
V=\left[\begin{array}{cc}
M*{I} & -{1 1}_{N, M} \\
-{1 1}_{N, M}^T & (N+M) {I}-{1 1}_{M, M}
\end{array}\right]
\label{eq:11}
\end{equation}
where the size of matrices $V_{11}$ and $B_{11}$ are $N \times N$, $V_{12}$ and $B_{12}$ being $N \times M$, $V_{22}$ and $B_{22}$ of $M \times M$. Now, for the majorization function $\tau(X,Z)$ as in \eqref{eq:5}, it is established that using SMACOF, the embeddings of the non-anchors can be learnt as a function of $X_A$ using the iterative solution given in equation \eqref{eq:12}:
\begin{equation}
{X_{NA}}^{(k)}={V}_{11}^{\dagger}\left({B}_{11} {X_{NA}}^{(k-1)}+\left({B}_{12}-{V}_{12}\right) {X}_A\right)
\label{eq:12}
\end{equation}
The proof of this formulation has been elaborated in \cite{di2017multidimensional}. Using $V_{12}$ and inverse of $V_{11}$ as in \eqref{eq:11}, the simplified update rule for computing the embeddings is given by:
\begin{equation}
{X_{NA}}^{(k)}=\frac{\left({B}_{11} {X_{NA}}^{(k-1)}+\left({B}_{12} + {1 1}_{N, M}\right) {X}_A\right)}{M}
\label{eq:13}
\end{equation}
However, in order to preserve the privacy of the user data, there is a restriction on the number of anchors $\left(M\right)$ used for computing the embeddings using \eqref{eq:13}. We provide the following Theorem that characterizes the connection between $M$ and the dimensionality $\left(d\right)$ of the data.\par

\textbf{Theorem 1}. To avoid learning exact feature embeddings in order to guarantee privacy, the number of anchors generated must be less than the original dimensionality of the data, i.e $M < d$.\par

\textbf{Proof}. Let there be $M$ anchors represented as $\left \{ A_1, A_2, \cdots , A_M \right \}$ and a non-anchor node $P$. Each of these nodes has features $\boldsymbol{x_{i}} \in \mathbb{R}^d$. Assuming that the exact distances between $P$ and corresponding anchors $\left( i.e \ d_{PA_i} \right)$ are known.  \par

The use of anchors for distributed sensor localization is a well-researched area \cite{khan2009distributed, khan2009diland}. In \cite{khan2009distributed}, the authors demonstrate that in Euclidean space, the minimum number of $\left(m+1\right)$ anchors with known locations is required to locate $N$ nodes (with unknown locations) in $\mathbb{R}^m$. The study also assumed that all non-anchor points must be within the convex hull of the anchors. However, even if non-anchors are placed in any location, at least $d$ anchors are needed to accurately recover their features in $\mathbb{R}^d$. Therefore, for the worst-case scenario, it can be concluded that having less than $d$ anchors, i.e. $M < d$, will not result in exact feature embeddings, providing privacy guarantees.\par

Next, the focus will be on exploring the use of these privacy-preserving embeddings to perform graph-based downstream tasks such as clustering.
\begin{algorithm}
 \caption{PPDA using MDS/Anchored-MDS}
 \begin{algorithmic}[1]
 \renewcommand{\algorithmicrequire}{\textbf{Input:}}
 \renewcommand{\algorithmicensure}{\textbf{Output:}}
 \REQUIRE M anchors ${X_{A}} \in \mathbb{R}^{M\times d}$, Incomplete Distance Matrix ${D}$, tolerance ($\epsilon$), maximum epochs ($t$), embedding dimension ($\widetilde{d}$)
 \ENSURE Embeddings $\left({X} \slash {X}_{NA}\right)$ in dimension $d$ or $\widetilde{d}$, Estimated user to user distance matrix ${K} \in \mathbb{R}^{N \times N}$
  
  \STATE Randomly initialize ${X^{(0)}} \slash {X_{NA}^{(0)}}  $ 
  \WHILE {$\sigma\left({X}^{(k-1)}\right)-\sigma\left({X}^{(k)}\right)<\epsilon$ or $k=t$} 
  \STATE Compute matrix $D$ using block matrices $D_{12}$ and $D_{22}$
  \STATE Compute matrix $B$
  \STATE Update ${X}$, ${X_{NA}}$ as in equation \eqref{eq:6}, \eqref{eq:13} respectively  
  \STATE Compute Stress $\left(\sigma \right)$ using \eqref{eq:1}
  \STATE Set $k \leftarrow k+1$
   \ENDWHILE
 \end{algorithmic} 
 \label{alg:PPDA}
 \end{algorithm}

\section{PPDA for Graph-based Clustering} \label{sec:clustering}

Graph-based clustering is a method of clustering data points into groups (or clusters) based on the relationships between the data points. This is typically done by constructing a graph as discussed in \hyperref[sec:graph learning]{3.2}. Clustering algorithms can then be applied to the graph to identify clusters of closely related data points. Graph-based clustering is often used in data mining and machine learning applications, as it can provide a more powerful and flexible way of identifying patterns and structures in data compared to traditional clustering methods. \par

In \cite{nie2016constrained}, Nie et al. proposed a new graph-based clustering method based on the rank constraint of Laplacian. The authors exploited an important property of the Laplacian matrix which tells that if $rank(\Theta) = N - k$, then the graph is an ideal representation for partitioning the data points into the $k$ clusters without using K-means or other discretization techniques that are often required in traditional graph-based clustering methods such as spectral clustering \cite{ng2001spectral}. Utilizing this characteristics, given the initial adjacency (A), they aim to learn a new adjacency matrix $P$ whose corresponding laplacian $\Theta_{P} = Q_{P} - \frac{P + P^T}{2}$ is constrained to rank $\left ( N - k \right )$. In addition, to prevent some rows of $P$ from being all zeros, they impose an additional constraint on $P$ to ensure that the sum of each row is equal to one. The solution to the optimization formulation as in \eqref{eq:17} is defined as Constrained Laplacian Rank (CLR) for graph clustering. 
\begin{equation}
\mathcal{L}_{\text {CLR}}=\min _{\sum_j p_{i j}=1, p_{i j} \geq 0, \operatorname{rank}\left(\Theta_P\right)=N-k}\|P-A\|_F^2
\label{eq:17}
\end{equation}

To build an initial graph, we solve the problem \eqref{eq:18} similar to \eqref{eq:15} which gives the estimate of $A$, utilizing privately estimated pairwise distance matrix $K$. 
\begin{equation}
\min _{\boldsymbol{a_i}^T \boldsymbol{1}=1, \boldsymbol{a_i} \geq \boldsymbol{0}, a_{i i}=0} \boldsymbol{a_{i}}^T\boldsymbol{k_{i}}+\gamma \left \| \boldsymbol{a_i} \right \|_2^2
\label{eq:18}
\end{equation}
In order to perform clustering based on probabilistic graphical models, the authors in \cite{JMLR:v21:19-276} introduced a general optimization framework for structured graph learning (SGL) enforcing eigenvalue constraints on graph matrix $\Theta$ as follows:
\begin{equation}
\begin{array}{ll}
\underset{\Theta}{\operatorname{maximize}} & \log \operatorname{det}(\Theta)-\operatorname{tr}(\Theta S)-\alpha h(\Theta) \\
\text { subject to } & \Theta \in \mathcal{S}_{\Theta}, \lambda(\mathcal{T}(\Theta)) \in \mathcal{S}_{\mathcal{T}}
\end{array}
\label{eq:25}
\end{equation}
where $\mathcal{S}_{\Theta}$ is the set of valid graph Laplacian matrix as in \eqref{eq:14} and $ \lambda(\mathcal{T}(\Theta)) $ denotes the eigenvalues of $\mathcal{T}(\Theta)$ with an increasing order, where $\mathcal{T}(\cdot)$ is a transformation on the matrix $\Theta$.
As demonstrated in Section \hyperref[sec: PPDA framework]{4}, we have the complete pairwise distance for each user estimated privately using PPDA and their corresponding embeddings as the algorithm's output, which can be used further to compute $S$.

\noindent Figure \ref{fig:pipeline} shows the overall pipeline for privacy preserved distributed graph learning and graph-based clustering. As we just need a adjacency matrix $A$ for clustering, which can be computed using pairwise distance matrix $K$, this framework can be adapted for any graph-based clustering algorithm beyond CLR, expanding its range of applications.

\section{Experiments} \label{sec:experiments}
In this section, we perform extensive experiments on widely adopted real as well as synthetic datasets. The efficacy of our proposed framework is elucidated by showing that structural properties are retained even without data sharing. We also evaluate graph-based clustering on datasets that are sensitive to privacy. Firstly, we introduce the experimental settings in section \hyperref[sec:expset]{6.1} and present the evaluation of the proposed framework in further subsections.
\begin{figure*}
     \centering
     \begin{subfigure}[b]{0.3\textwidth}
         \centering
         \includegraphics[width=\textwidth]{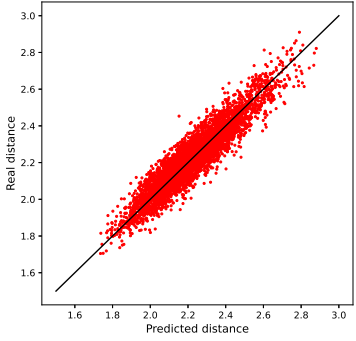}
         \caption{ER with $\boldsymbol{N = 100}$ and $\boldsymbol{M = 199}$}
         \label{fig:ER}
     \end{subfigure}
     \hfill
     \begin{subfigure}[b]{0.3\textwidth}
         \centering
         \includegraphics[width=\textwidth]{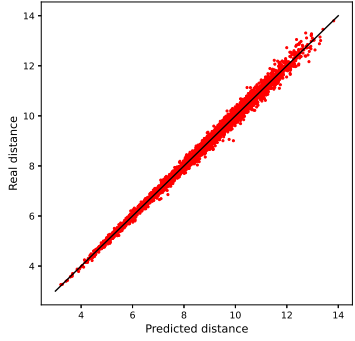}
         \caption{BA with $\boldsymbol{N = 100}$ and $\boldsymbol{M = 399}$}
         \label{fig:BA}
     \end{subfigure}
     \hfill
     \begin{subfigure}[b]{0.3\textwidth}
         \centering
         \includegraphics[width=\textwidth]{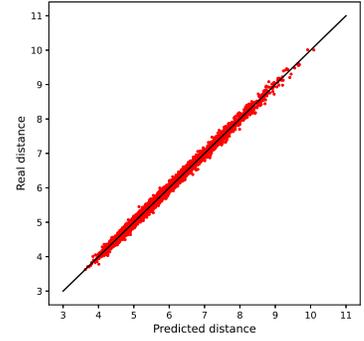}
         \caption{RGG with $\boldsymbol{N = 100}$ and $\boldsymbol{M = 699}$}
         \label{fig:RGG}
     \end{subfigure}
        \caption{True euclidean distance (vertical axis) vs. predicted euclidean distance (horizontal axis) for PPDA}
        \label{fig:distance_plots}
\end{figure*}

\subsection{Experimental setting} \label{sec:expset}
For conducting experiments involving synthetic data, we create various datasets based on diverse graph structures $\mathcal{G}$. We utilize three synthetic graphs with parameters $N,M,m,p$ namely: (i) \textit{Erdos Renyi (ER)} graph with $\left(N + M\right)$ nodes and $p = 0.1$ (ii) \textit{Barabasi Albert (BA)} graph with $\left(N + M\right)$ nodes and $m = 2$ and (iii) \textit{Random Geometric (RGG)} graph with $\left(N + M\right)$ nodes and $p = 0.1$, where $N$ is the number of non-anchors and $M$ is the number of anchors. Initially, we create an IGMRF model using the true precision matrix $\Theta_{\mathcal{G}}$ as its parameter, which complies with the Laplacian constraints in \eqref{eq:14} and the above graph structures. Than, we draw total of $d$ samples $\left\{\mathbf{x}_i \in \mathbb{R}^{N+M}\right\}_{i=1}^d$ from the IGMRF model as $\mathbf{x}_i \sim \mathcal{N}\left(0, \Theta_{\mathcal{G}}^{\dagger}\right)$. We sample $M$ nodes from this same distribution to use as anchors.
 We use the metrics namely relative distance error $\left(RE \right)$ for distance approximation and F-score $\left(FS \right)$ to manifest neighborhood structure preservation. These performance measures are defined as:
\begin{equation}
\text { Relative Error }=\frac{\left\|\hat{D}-D_{\text {true }}\right\|_F}{\left\|D_{\text {true }}\right\|_F}
\end{equation}
\begin{equation}
\text { F-Score }=\frac{2 \mathrm{tp}}{2 \mathrm{tp}+\mathrm{fp}+\mathrm{fn}}
\end{equation}
where $\hat{D}$ is the estimated distance matrix based on the PPDA algorithm and $D_{\text{true}}$ is the original distance matrix. For each node, we consider their $k$ nearest neighbors. True positive (tp) stands for the case when one of the other nodes is the actual neighbor and the algorithm detects it; false positive (fp) stands for the case when the other node was not a neighbor but the algorithm detects so; and false negative (fn) stands for the case when the algorithm failed to detect an actual neighbor. The F-score takes values in [0, 1] where 1 indicates perfect neighborhood structure recovery \cite{egilmez2017graph}.

\subsection{Experiments for PPDA} \label{sec:ppgl}

\subsubsection{Synthetic Datasets:} \label{sec: synthetic dataset}
For learning a graph structure, we need the pairwise distance matrix or sample covariance matrix as discussed in section \hyperref[sec:graph learning]{3.2} and \hyperref[sec:PGM]{3.3}. However, since we are not sharing the features, we estimate these parameters through PPDA and then learn the graph structure. It's worth noting that we are not recovering the exact features but we ensure to preserve the network structure.
\noindent For this experiment, we use graph structures and samples generated as described in section \hyperref[sec:expset]{6.1}. We examine two cases for learning the embeddings: $\left(i\right)$ in the original dimension and $\left(ii\right)$ in a lower dimension. Table \ref{table:synthetic_graph1} summarizes the performance of PPDA for various numbers of non-anchors and anchors for the case $\left(i\right)$. We consider various dimensions of data when generating samples for each synthetic graph to demonstrate the effectiveness of the algorithm across a broad range of dimensionality. For ER we fix $d = 200$, $d = 400$ for BA, and $d = 700$ for RGG. As stated in Theorem $1$, the number of anchors $\left(M\right)$ chosen for each dataset is $(d - 1)$. For case $\left(i\right)$ to learn ${X_{NA}}$, we run the Algorithm \ref{alg:PPDA} for tolerance $\epsilon = 0.001$ and maximum epochs $t = 5000$. Note that, for computing F-score, we are considering $k = 10$ nearest neighbors. It can be observed from table \ref{table:synthetic_graph1} that for $M >> N$ the relative distance error is significantly low, and the F-score is high, indicating that structural information is well preserved. The plot in Figure \ref{fig:distance_plots} illustrates that the actual and estimated distance pairs are situated closely around the line $y = x$ (shown in black), indicating that PPDA provides accurate distance approximation. \par
\begin{figure}
\begin{subfigure}{.46\textwidth}
  \centering
  \includegraphics[width=0.48\linewidth]{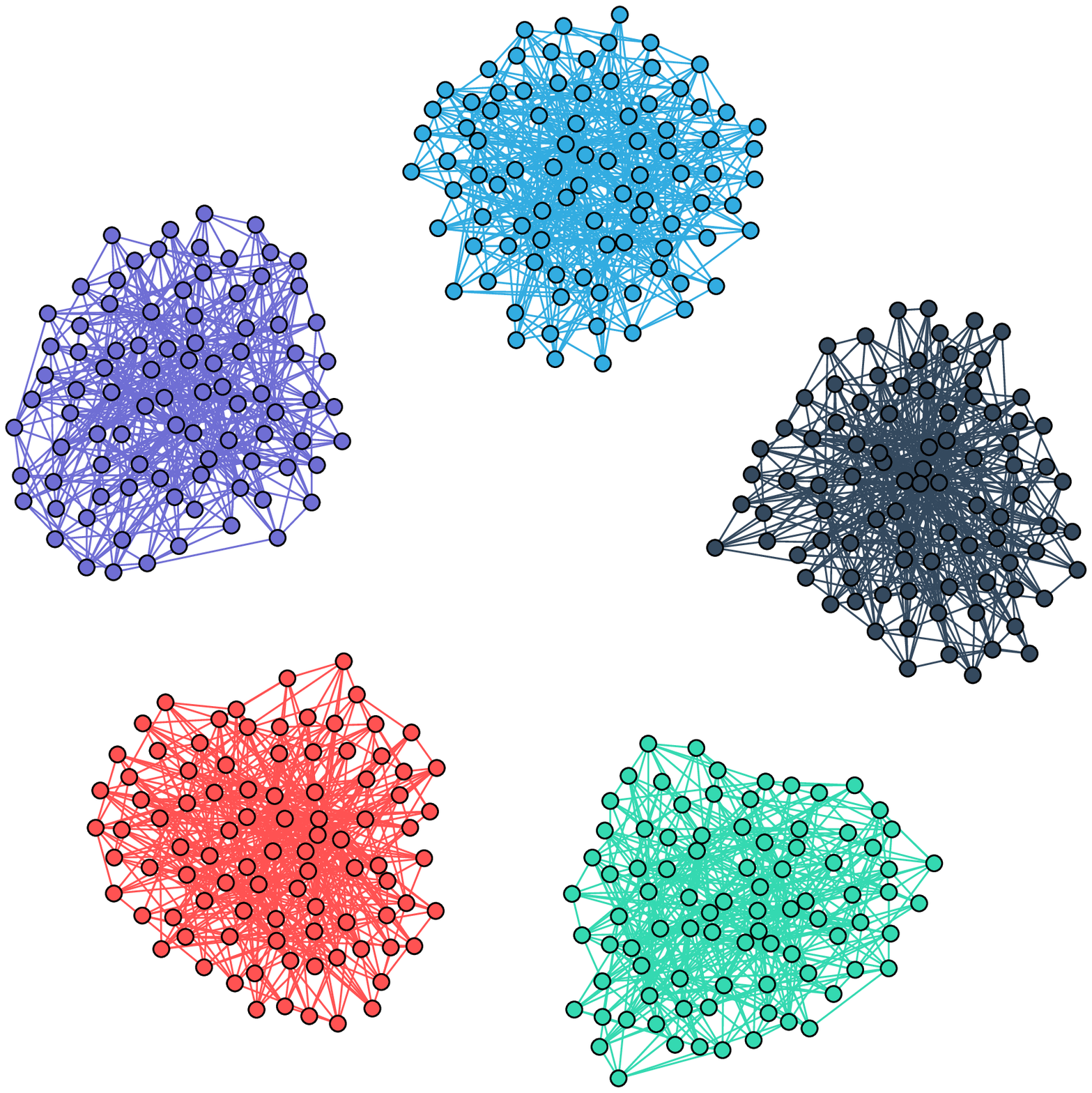}
  \includegraphics[width= 0.48\linewidth]{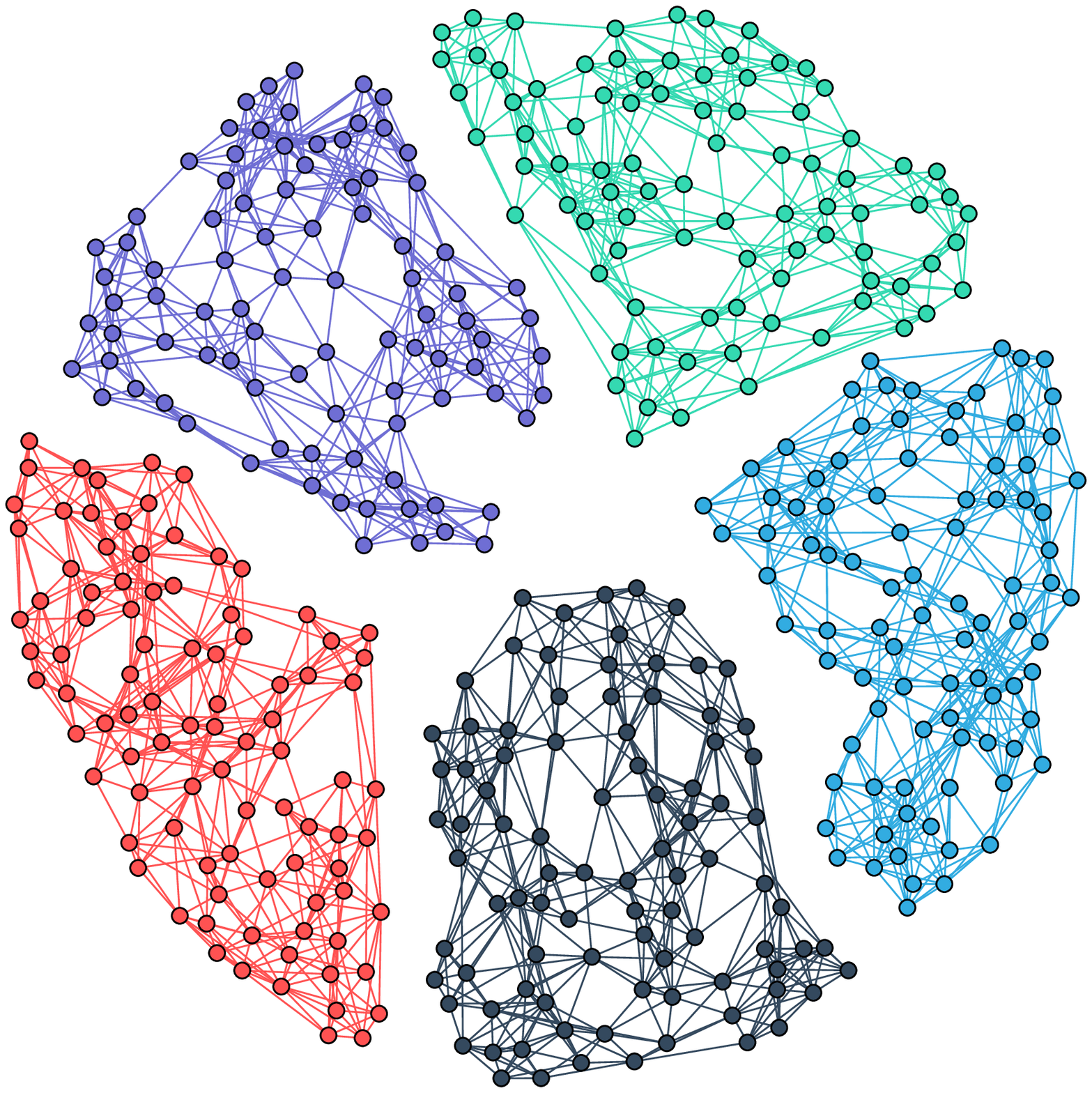}
\end{subfigure}
\vspace{-1.5em}
\caption{The image shows the clustered graph estimated using SGL for synthetic dataset with 5 clusters with $\widetilde{d} = 3$ for (a) original features (Left) (b) without knowing features (Right)}
\label{fig: synthetic_graph}
\end{figure}

\begin{table}
\begin{tabular}{|| m{1.4cm} m{1.3cm} m{1.3cm}  m{1.2cm} m{1.1 cm} ||} 
 \hline
 \textbf{Datasets} & \textbf{Nodes (N)} & \textbf{Anchors (M)} & \textbf{Error (RE)} & \textbf{F-score} \\ [0.5ex]
 \hline \hline
\multirow{3}{4em}{ER} & 100 & 199 & 0.0292 &   0.7503 \\ 
& 500 & 199 & 0.0264  & 0.6192 \\ 
& 1000 & 199 & 0.0258 & 0.5773\\
 \hline
\multirow{3}{4em}{BA} & 100 & 399 &  0.0157 & 0.9406 \\ 
& 500 & 399 & 0.0167 & 0.9147 \\ 
& 1000 & 399 & 0.0174  & 0.8894\\
 \hline
 \multirow{3}{4em}{RGG} & 100 & 699 & 0.0127 & 0.9117 \\ 
& 500 & 699 & 0.0140 &  0.8200 \\ 
& 1000 & 699 & 0.0148 &  0.7220 \\[1 ex]
 \hline
 \end{tabular}
 \vspace*{2mm}
 \caption{Results of distance approximation error (RE) and F-score for embeddings in the original dimension}

\label{table:synthetic_graph1}
\end{table}
For case $\left(ii\right)$, we generate isotropic gaussian blobs to perform clustering. We created 100 samples per cluster for a total of 5 clusters with $d = 1000$, where $d$ is the number of features for each sample. We divided this dataset by keeping $10\%$ of the samples as anchors and the rest as non-anchors. The embeddings $X$ are then learned in a reduced dimension with $\widetilde{d} = 3$. Despite the reduced dimension, PPDA is able to approximate the distances with a relative distance error of 0.1805. Furthermore, utilizing the SGL algorithm described in Section \hyperref[sec:clustering]{5}, we construct a $k$-component graph with $k = 5$ using the sample covariance matrix for original features as well as the learned embeddings. From Figure \ref{fig: synthetic_graph}, through visual inspection, it can be observed that even without feature sharing we are able to cluster the nodes exactly similar to one with original features, which is notable.

\begin{center}
\begin{table}
\begin{tabular}{||m{1.24cm} m{0.85cm} m{1.4cm} m{1.5cm}  m{1.5 cm}||} 
 \hline
\textbf{Datasets} & \textbf{Nodes (N)} & \textbf{RE/FS  \hspace{5em} \hfill (r = 0.1)} & \textbf{RE/FS  \hspace{5em} (r = 0.3)} & \textbf{RE/FS  \hspace{5em} (r = 0.5) }\\ [0.5ex] 
 \hline\hline
\multirow{3}{4em}{ER} & 100 & 0.025/0.789 &  0.021/0.831  & 0.017/0.878 \\ 
& 500 & 0.022/0.690  & 0.017/0.784  & 0.012/0.865 \\ 

 \hline
\multirow{3}{4em}{BA} & 100 & 0.014/0.948 & 0.012/0.957  &    0.010/ 0.961 \\ 
& 500 &  0.015/0.925  & 0.013/0.934 & 0.011/0.952 \\ 

 \hline
 \multirow{3}{4em}{RGG} & 100 &  0.012/0.920  & 0.010/0.932 &  0.009/0.949\\ 
& 500 &0.013/0.834 & 0.011/0.864 & 0.010/0.898 \\ [1 ex]
 \hline
 \end{tabular}
  \vspace*{2mm}
 \caption{Distance approximation with partial inter-user distances known}
 \label{table:synthetic_graph3}
\end{table}
\end{center}
\vspace{-1em}
\subsubsection{Practical use case:}

 For many real-world applications across various domains, such as financial institutes, electronic health record mining, medical data segmentation, and pharmaceutical discovery, sharing data within an institution is permitted. This allows us to have known block entries in the fully incomplete distance matrix $K$. Let us denote the set of known blocks as $\left\{S_1, S_2, \cdots, S_R\right\}$. Hence for the matrix $K$, $K_{ij}$ is known $\forall \left \{ \boldsymbol{x_i}, \boldsymbol{x_j} \right \}_{i\neq j} \in S_r$, where $r \in \left\{1, 2, \cdots, R\right\}$. To illustrate this use case, we assume that a certain percentage of total entries in the distance matrix for $N$ nodes is available. For the same set of non-anchors and anchors used in case $\left(i\right)$ of Section \hyperref[sec: synthetic dataset]{6.2.1}, we examine various ratios $r$ between the known and overall entries of the distance matrix. From Table \ref{table:synthetic_graph3}, it is clearly evident that as the number of known entries increases, there is a decrease in relative distance error (RE) and an increase in the corresponding F-score.

\subsection{PPDA for Private Graph-based Clustering} \label{sec:graph_clustering}
We conduct experiments on the widely adopted benchmark datasets to showcase the performance of private distributed graph learning and graph-based clustering. For unsupervised data clustering, we explore the application of multiscale community detection on similarity graphs generated from data. Specifically, we focus on how to construct graphs that accurately reflect the dataset's structure in order to be employed inside a multiscale graph-based clustering framework. We have evaluated our framework on 8 benchmark datasets from the UCI repository (Table \ref{table:real_dataset} summarizes dataset statistics) \cite{Dua:2019}. All datasets have ground truth labels, which we use to validate the results of the clustering algorithm.\par
\begin{center}
\begin{table}
\begin{tabular}{||c c c c||} 
 \hline
 \textbf{Datasets} & \textbf{Nodes (N)} & \textbf{Dimension (d)} & \textbf{Classes (c)} \\ [0.5ex] 
 \hline\hline
 Iris & 150 & 4 & 3 \\ 
 \hline
 Glass & 214 & 9 & 6 \\
 \hline
 Wine & 178 & 13 & 3 \\
 \hline
 Control Chart & 600 & 60 & 6 \\
 \hline
 Parkinsons & 195 & 22 & 2 \\
 \hline
  Vertebral & 310 & 6 & 3 \\
 \hline
  Breast tissue & 106 & 9 & 6 \\
 \hline
 Seeds & 210 & 7 & 3 \\ [1ex] 
 \hline
\end{tabular}
 \vspace*{1em}
\caption{Dataset Statistics for 8 benchmark datasets from UCI repository}
 \label{table:real_dataset}
 \end{table}
\end{center}

\begin{figure}
\begin{subfigure}{.46\textwidth}
  \centering
  \includegraphics[width=.49\linewidth]{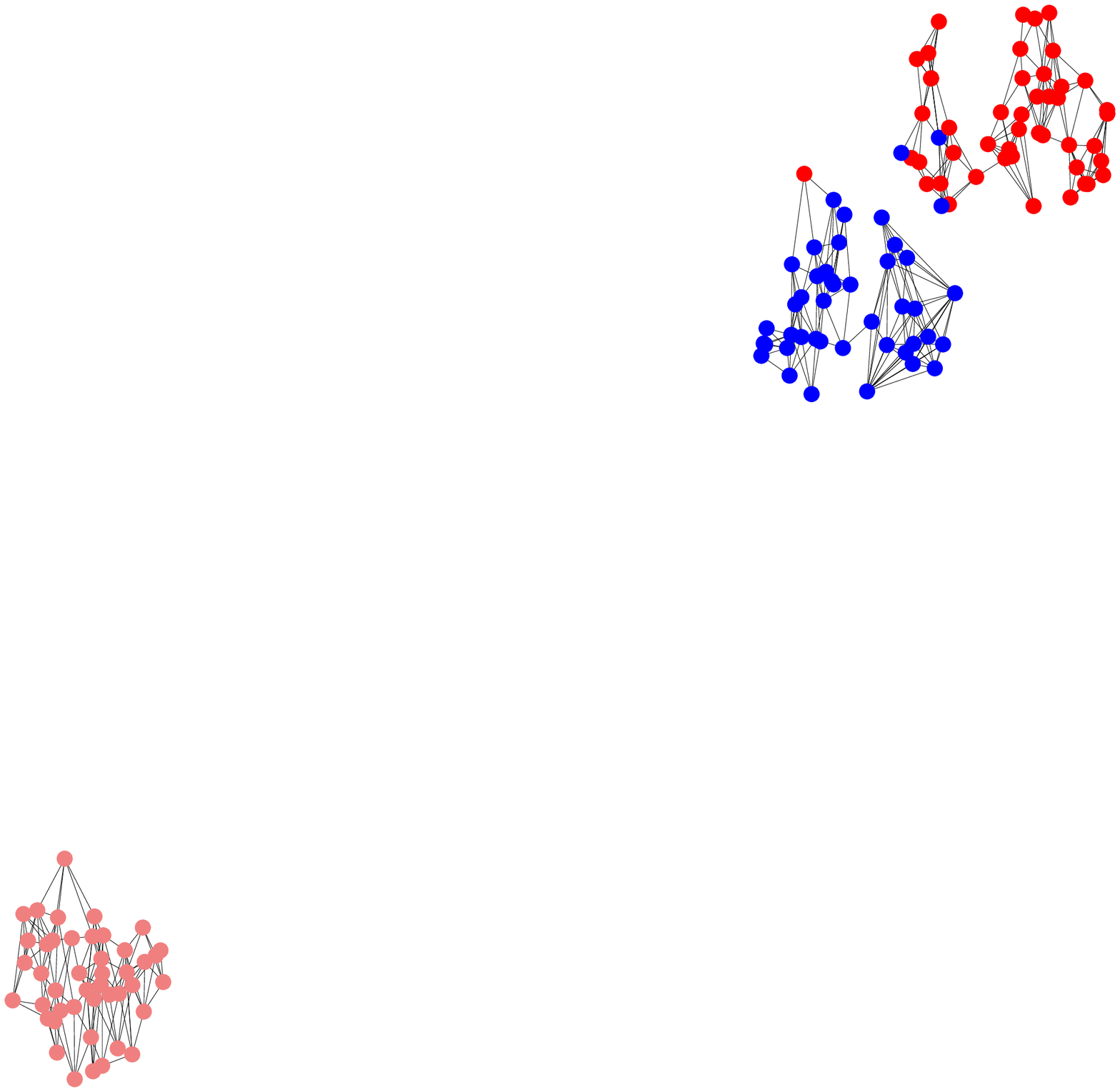}
  \includegraphics[width=.49\linewidth]{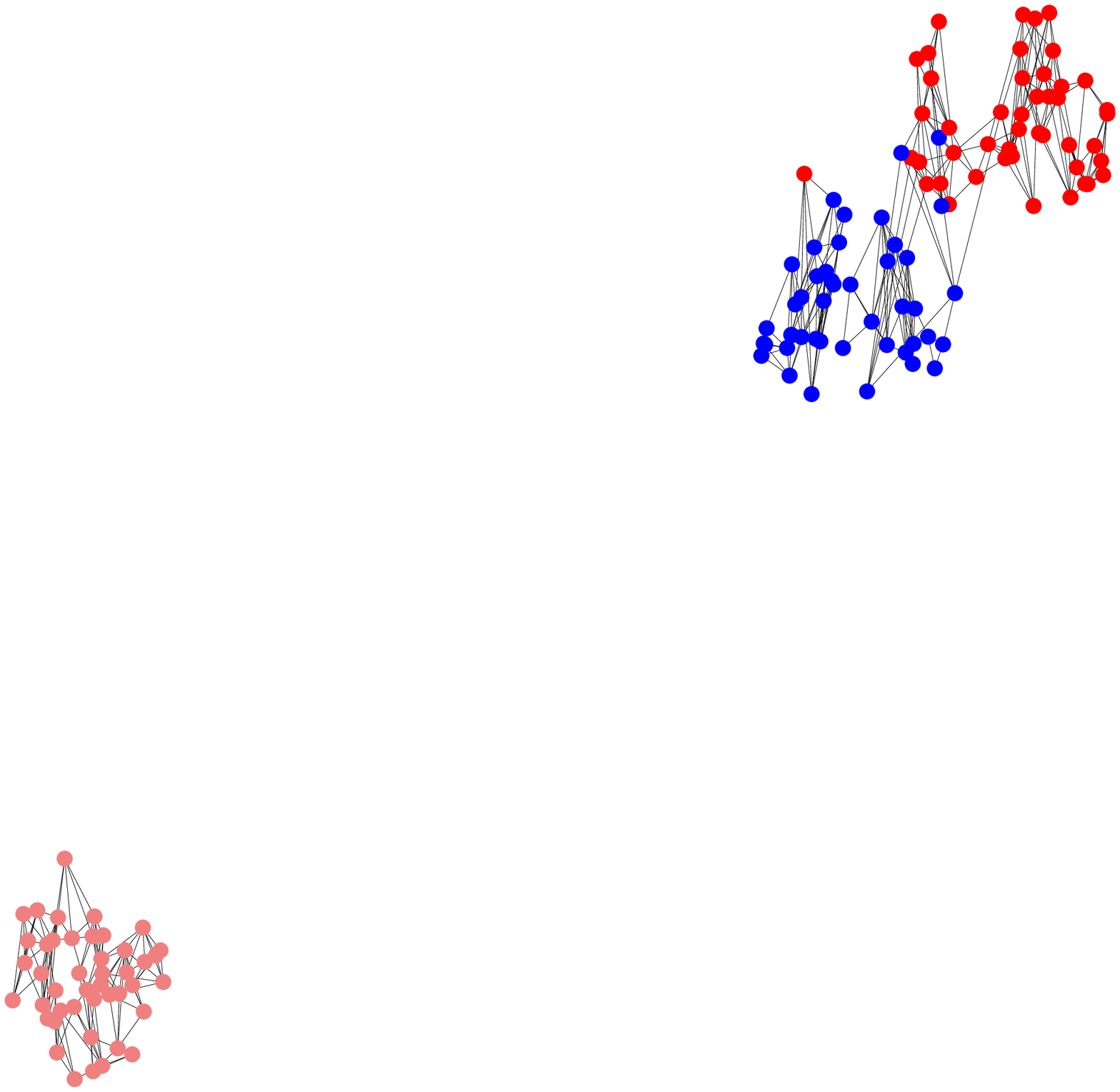}
  \label{fig:sfig1}
\end{subfigure}
\vspace{-2em}
\begin{subfigure}{.46\textwidth}
  \centering
  \includegraphics[width=.49\linewidth]{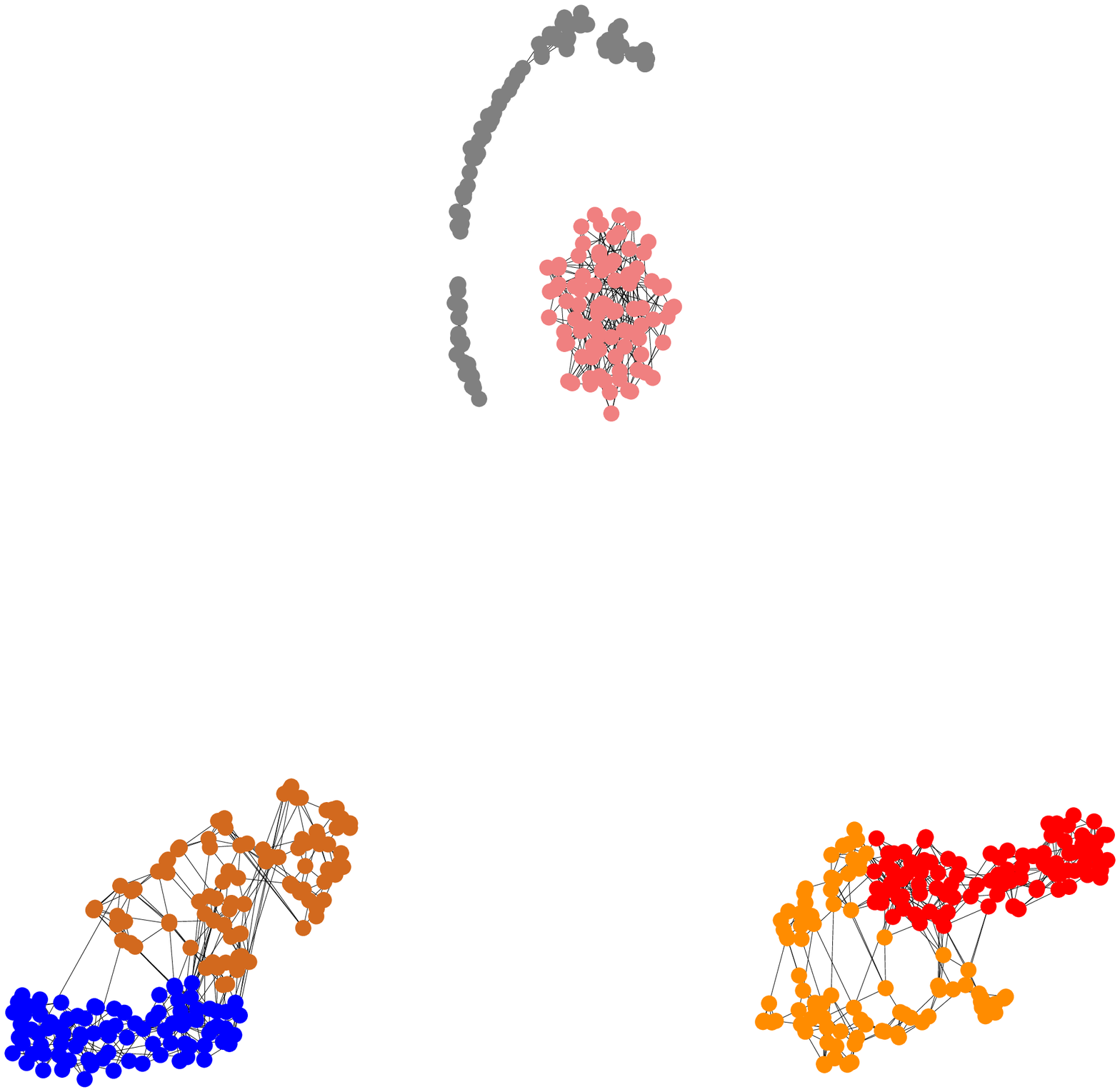}
   \includegraphics[width=.49\linewidth]{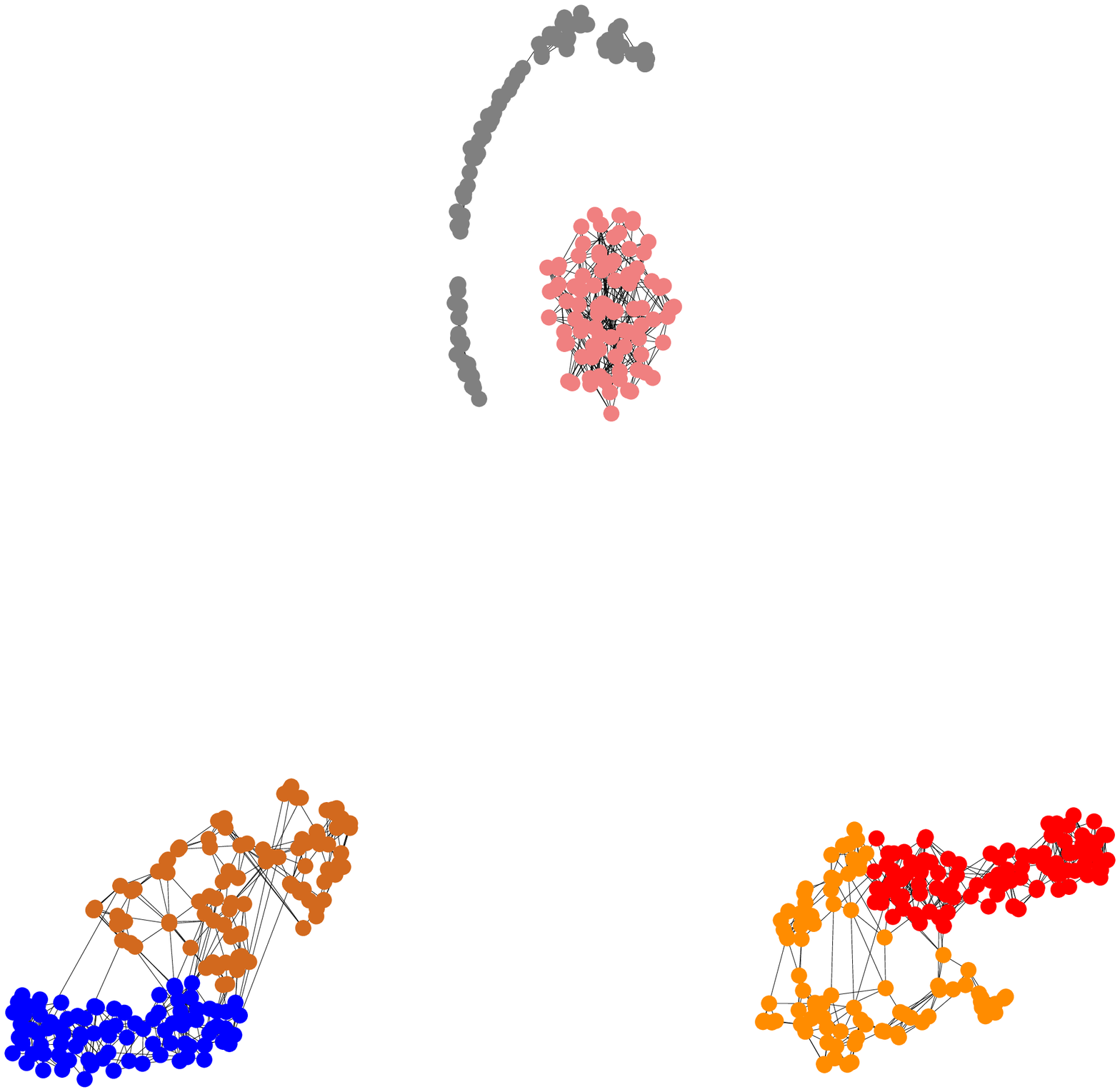}
  \label{fig:sfig2}
\end{subfigure}
\vspace{-1em}
 \begin{subfigure}{0.23\textwidth}
         \centering
         \includegraphics[width=\textwidth]{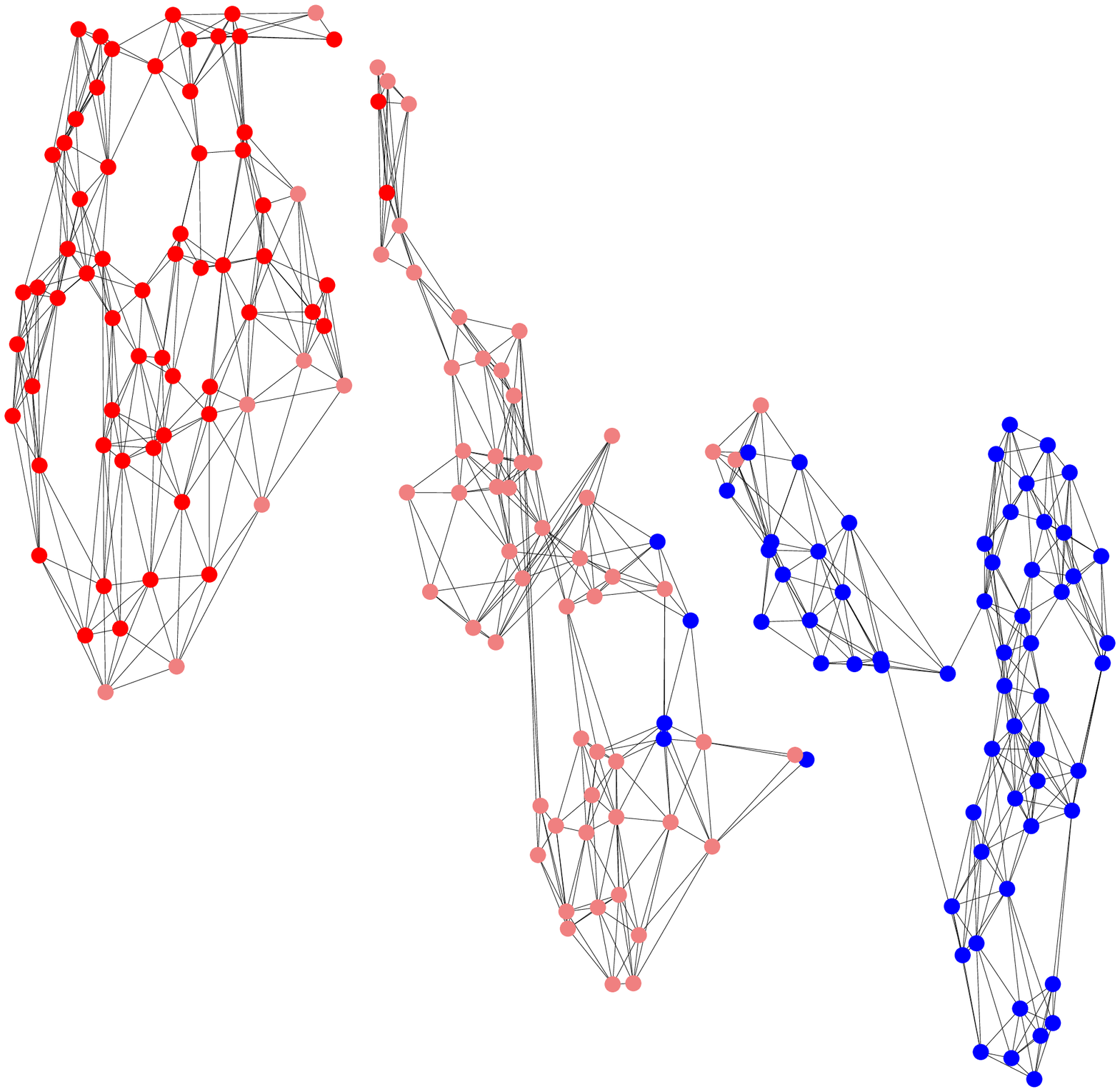}
         
         \caption{Non-Private}
     \end{subfigure}
     \begin{subfigure}{0.23\textwidth}
         \centering
         \includegraphics[width=\textwidth]{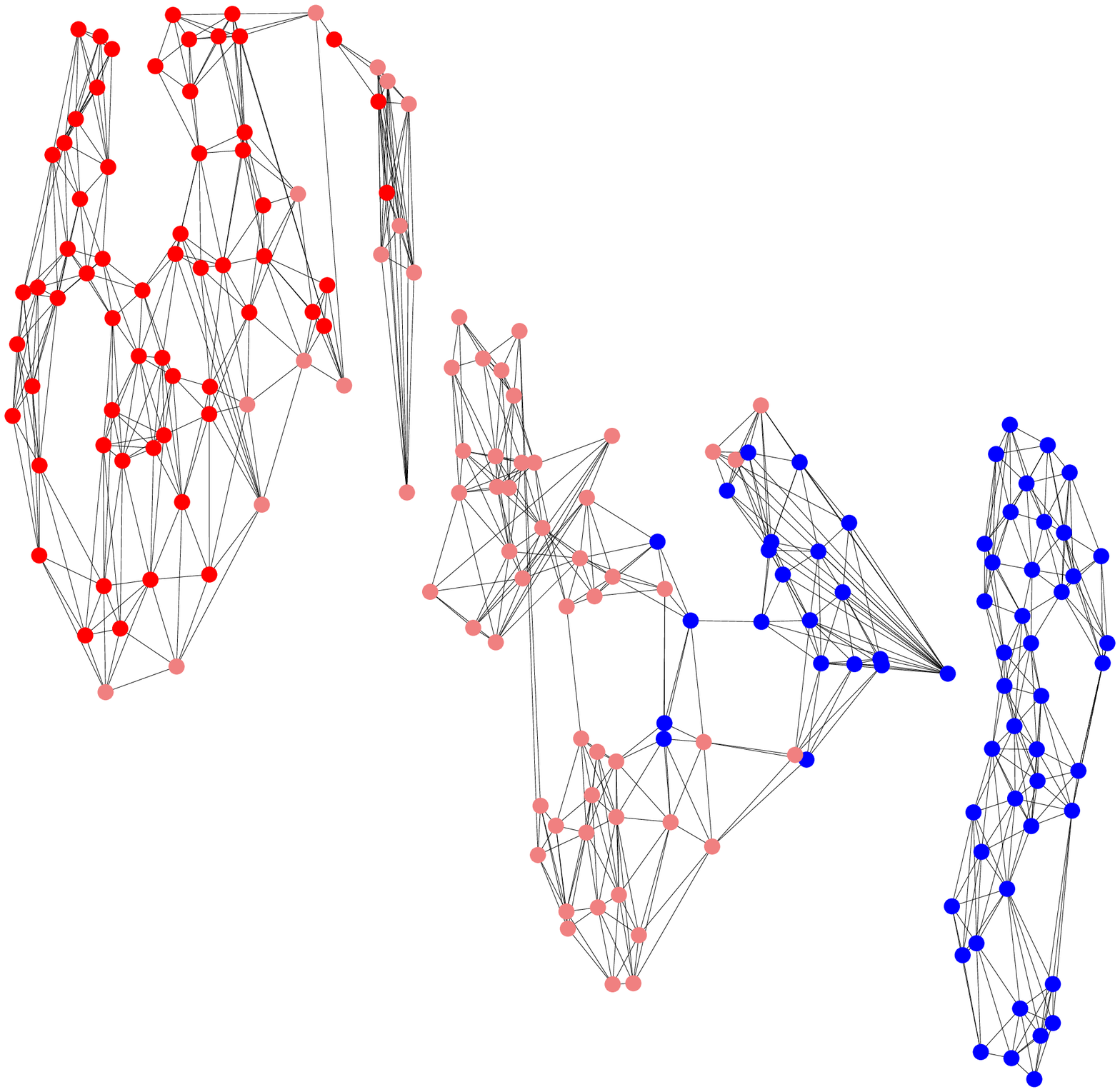}
        
         \caption{Private}

     \end{subfigure}

\caption{Graph Clustering performed using CLR for Iris (Top), Control Chart (Middle), Seeds (Bottom) Dataset with and without feature sharing}
\label{fig:real_graphs}
\end{figure}

\vspace{-2em}
\textbf{Experimental Details:} For every dataset, we assume that each sample belongs to one user in a distributed setting. As it is more efficient to sample anchors from a similar distribution as of original data, we split the given datasets keeping $10\%$ of samples as anchors and rest as non-anchors. With only $(d-1)$ anchors, we are able to preserve neighborhood structure ensuring privacy, as is evident from the F-score presented in Table \ref{table:performance_metrics}. We then employ the Constrained Laplacian Rank (CLR) algorithm for graph-based clustering. To construct the initial graph from the data we estimate the Adjacency matrix by solving the optimization formulation given in equation \eqref{eq:18} as discussed in Section \hyperref[sec:clustering]{5}. To validate the graph construction under clustering, we compute Normalised Mutual Information (NMI) \cite{strehl2002cluster} and Adjusted Rand Index (ARI) \cite{hubert1985comparing} as quality metrics.
Results obtained with attribute sharing are considered benchmarks. To set the number of clusters, we use the number of classes in the ground truth (c) as input to the CLR algorithm. Table \ref{table:performance_metrics} provides strong evidence that the validation metrics are comparable whether the original data was shared or the attributes were kept private. These results establish a strong foundation for privacy-preserving graph-based clustering. Figure \ref{fig:real_graphs} shows the graph constructed along with clusters using raw data and with our proposed framework. It can be seen that the clustering performance is almost the same for both cases, which is remarkable.

\begin{center}
\begin{table}[h]
\begin{tabular}{|| m{1.4cm} |  m{0.8cm}| m{0.98cm} | m{0.98cm} | m{0.98 cm} |m{0.98cm}|| } 
 \hline
\textbf{Datasets} & \textbf{F-score} &\multicolumn{2}{|c|}{\textbf{NMI}} & \multicolumn{2}{|c|}{\textbf{ARI}}\\ [0.5ex] 
 \hline
  & & \textbf{Non-Private} & \textbf{Private} & \textbf{Non-Private} & \textbf{Private} \\ [0.5ex]
 \hline
 Iris & 0.8659 &0.892 & 0.785 & 0.913 & 0.718\\ 
 \hline
 Glass & 0.9337 &0.354 & 0.341 & 0.199 & 0.173 \\
 \hline
 Wine & 0.9934 &0.377 & 0.377 & 0.381 & 0.381\\
 \hline
 Control Chart &  0.8271 & 0.811 & 0.811 & 0.620 & 0.620 \\
 \hline
 Parkinsons & 0.9880 & 0.016 & 0.016 & 0.055 & 0.055 \\
 \hline
  Vertebral & 0.7612 & 0.468 & 0.459 & 0.338 & 0.331 \\
 \hline
  Breast tissue & 0.9938 & 0.341 & 0.341 & 0.162 & 0.162 \\
 \hline
 Seeds & 0.9745 & 0.701 & 0.619 & 0.739 & 0.549 \\ [1ex] 
 \hline
\end{tabular}
\vspace*{2mm}
\caption{Performance evaluation of Graph-based clustering with and without data sharing}
 \vspace{-10mm}
 \label{table:performance_metrics}
 \end{table}
\end{center}

\subsection{Experiment on Sensitive Benchmark Data}
\begin{figure}
     \centering
     \begin{subfigure}[b]{0.23\textwidth}
         \centering
         \includegraphics[width=\textwidth]{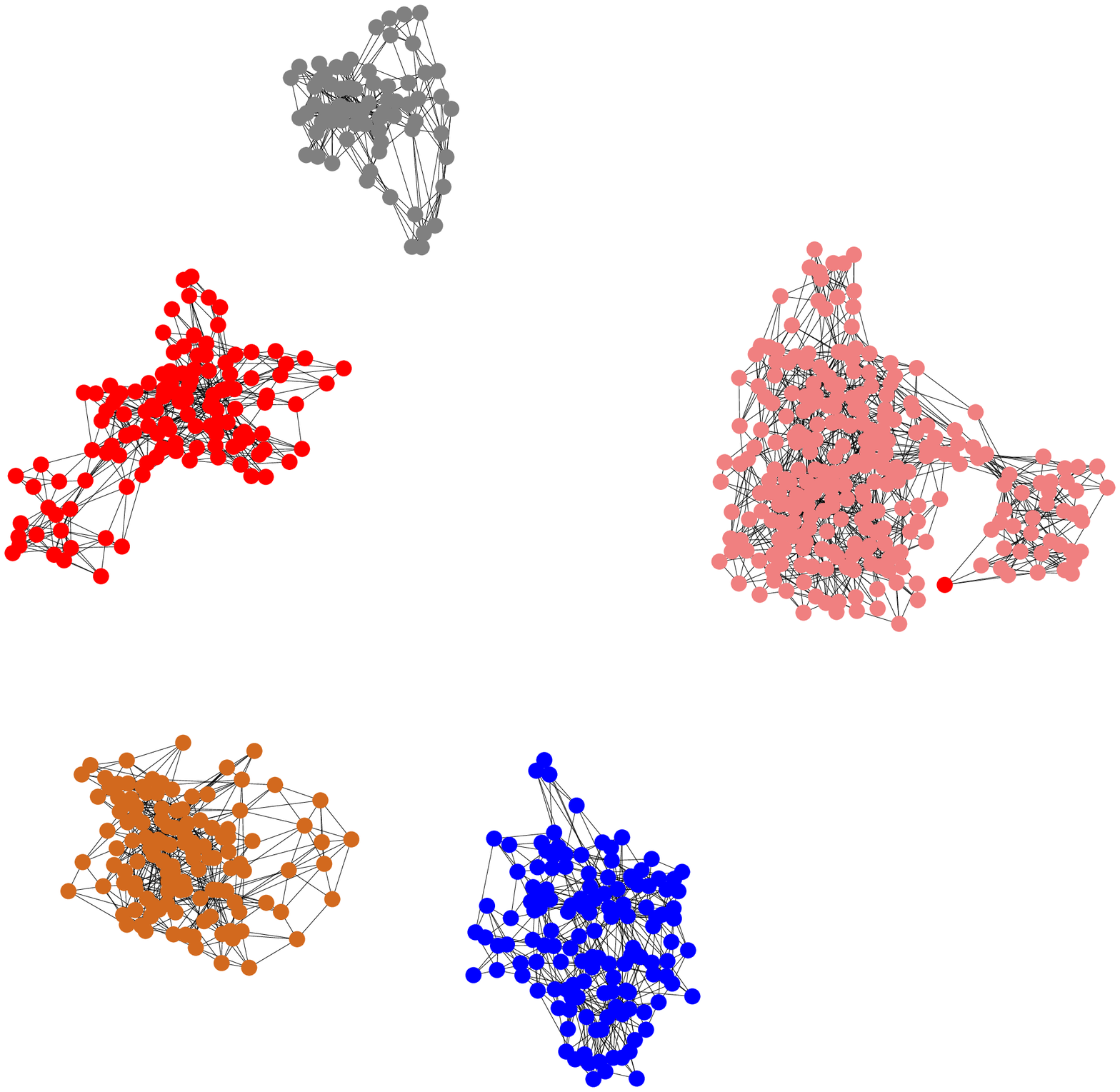}
         \vspace{-1em}
         \caption{Non-Private}
     \end{subfigure}
     \hfill
     \begin{subfigure}[b]{0.23\textwidth}
         \centering
         \includegraphics[width=\textwidth]{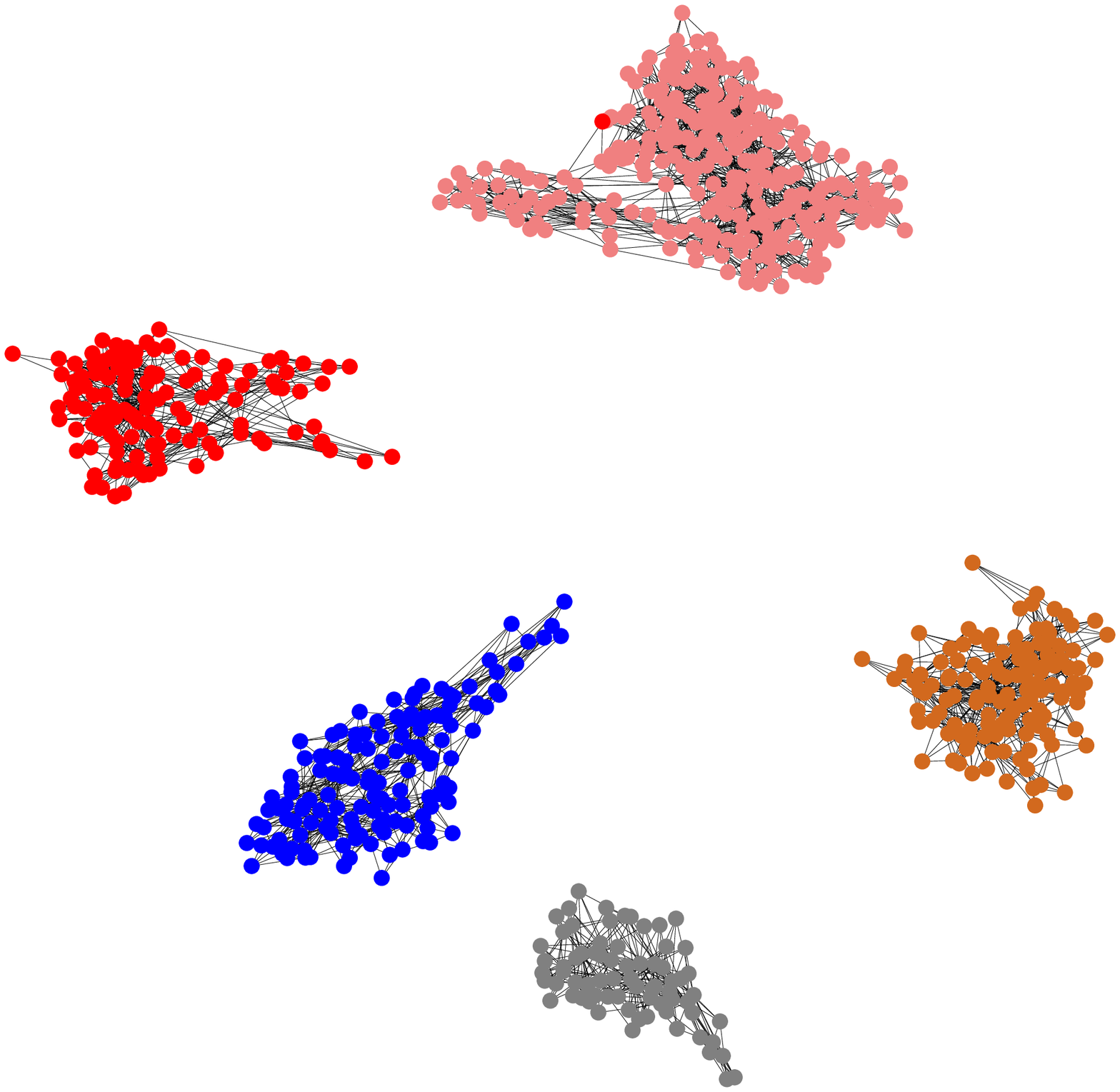}
         \vspace{-1em}
         \caption{Private}

     \end{subfigure}

        \caption{The image shows the clustered graph estimated using CLR for the PANCAN dataset with the following metrics for non-private/private graphs: NMI = 0.9943/0.9841 and ARI = 0.9958/0.9875}
        \label{fig:PANCAN}
    \vspace{-1em}
\end{figure}

\noindent Additionally we also evaluated our framework on a privacy-critical medical dataset. We consider the RNA-Seq Cancer Genome Atlas Research Network dataset (PANCAN) \cite{weinstein2013cancer} available at UCI repository \cite{Dua:2019}. This dataset is a random extraction of gene expressions of patients having different types of tumor namely: 
breast carcinoma (BRCA), kidney renal clear-cell carcinoma (KIRC), lung adenocarcinoma (LUAD), colon adenocarcinoma (COAD), and prostate adenocarcinoma (PRAD). The data set consists of $801$ labeled samples with $20531$ genetic features. This dataset's objective is to group the nodes into tumor-specific clusters based on their genetic characteristics. We take $10\%$ of the total samples as anchors and use Algorithm \ref{alg:PPDA} to learn the embeddings with $\widetilde{d} = 500$. Figure \ref{fig:PANCAN} shows the clustered graph built using CLR with those cancer types being labeled with colors lightcoral, gray, blue, red, and chocolate respectively. The clustering performance is highly comparable with NMI being 0.9943 and 0.9841, ARI being 0.9958 and 0.9875 for shared and private data respectively.  

\subsection{Further Analysis:}
\begin{figure}
     \centering
     \begin{subfigure}[b]{0.23\textwidth}
         \centering
         \includegraphics[width=\textwidth]{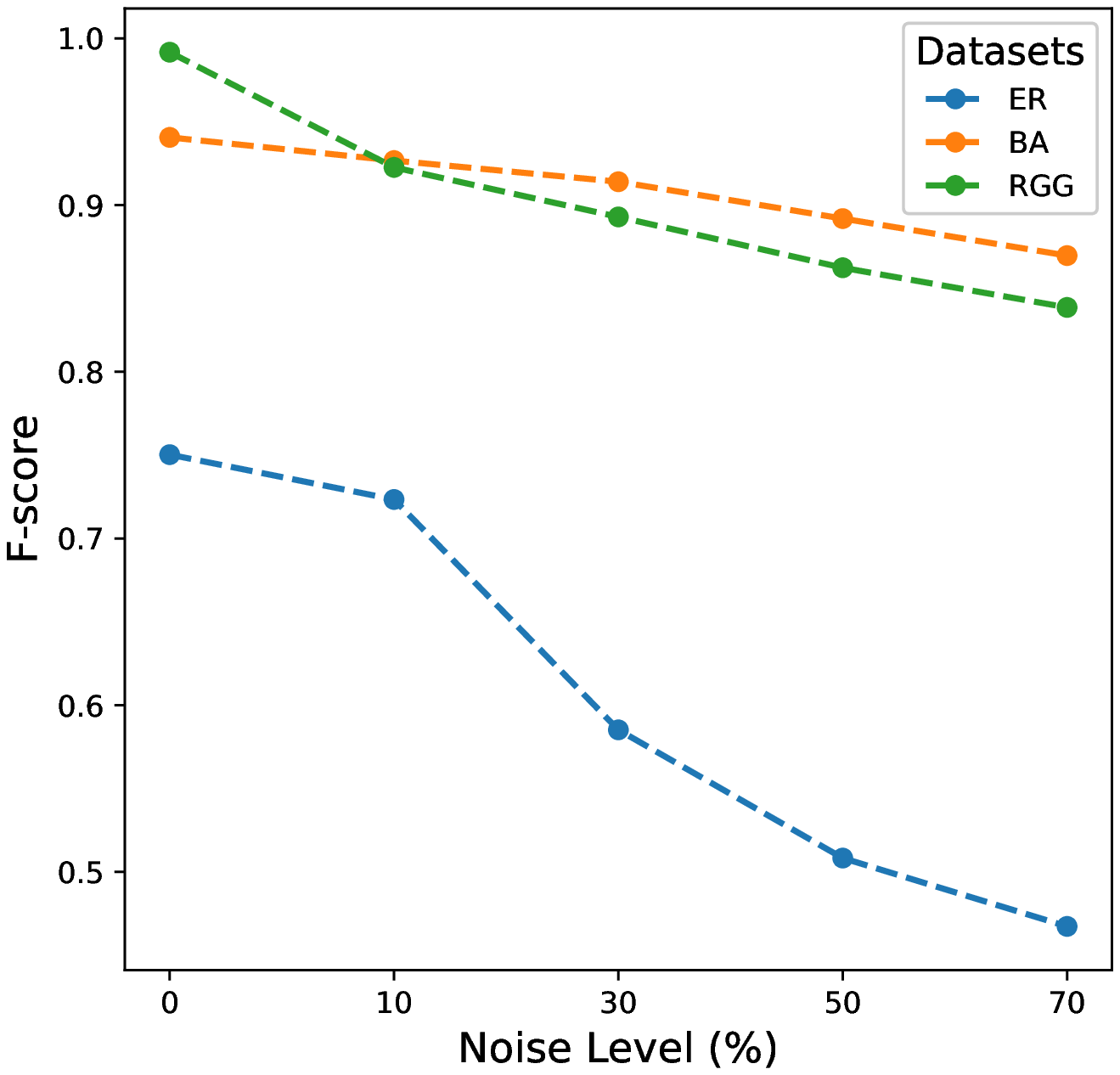}
        
         \label{fig:F-score noise}
     \end{subfigure}
     \hfill
     \begin{subfigure}[b]{0.23\textwidth}
         \centering
         \includegraphics[width=\textwidth]{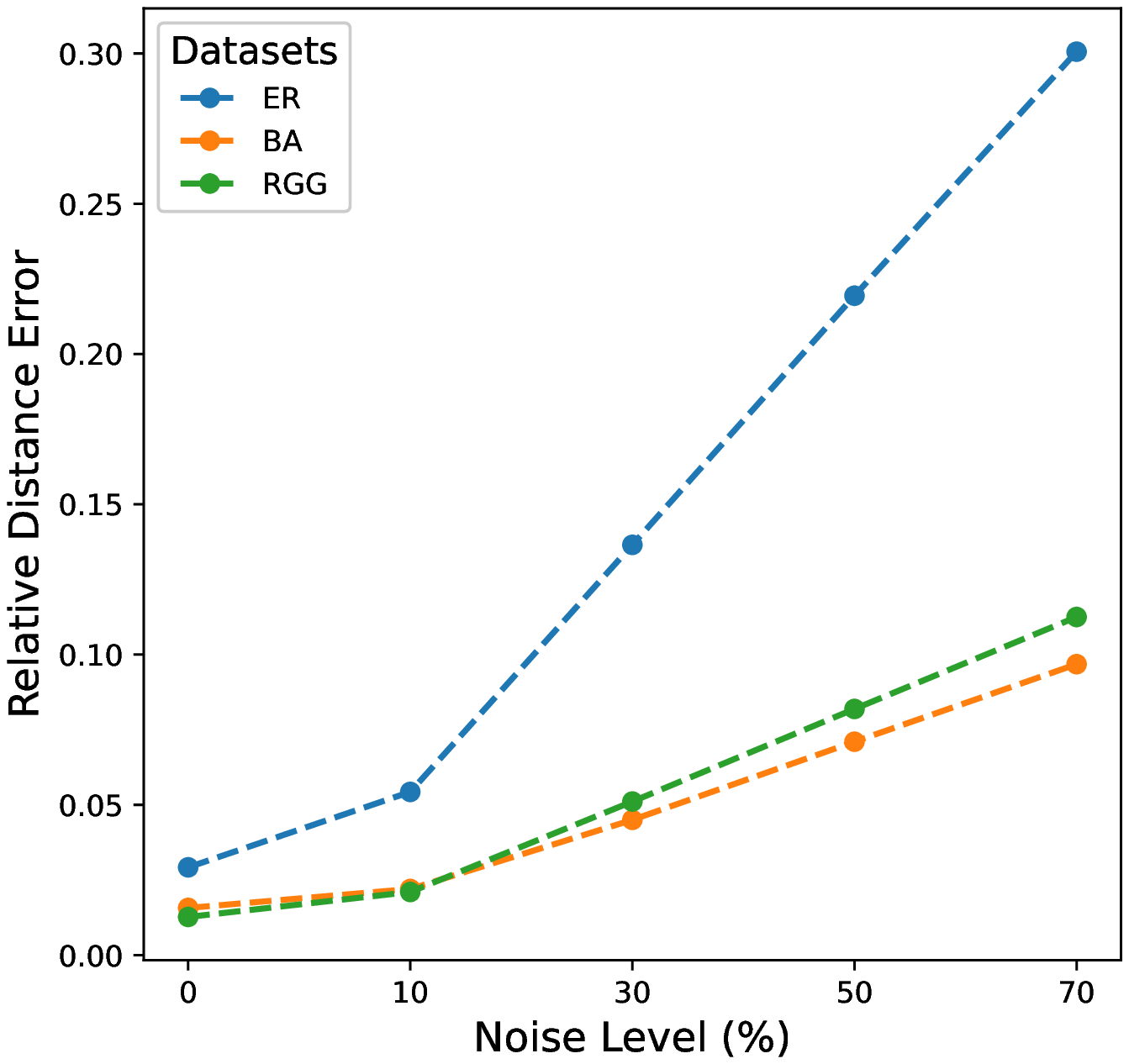}
        
         \label{fig:RDE Noise}
     \end{subfigure}
  \vspace{-1.5em}
        \caption{Analysis of performance metrics for (a) F-score (Left) and (b) Relative Distance Error (Right) with noise-induced distance parameters for three synthetic datasets namely ER, BA, and RGG with no noise and with a noise level of $\boldsymbol{10\%}$, $\boldsymbol{30\%}$, $\boldsymbol{50\%}$ and $\boldsymbol{70\%}$. }
        \label{fig:noise_plots}
\end{figure}

\textbf{Noisy Parameters:} Various factors can introduce noise in the distance measurements while sending it to the server, which may impact the effectiveness of the system. Therefore, it is necessary to investigate the impact of noise on the performance of the proposed framework. To accomplish this, random noise uniformly distributed in the range $\left[0, c\right]$ is added to each client to anchor distance measurements that are to be sent to the server. Here $c$ is set as $0.1, 0.3, 0.5$, and $0.7$ respectively to examine different levels of noise. As shown in Figure \ref{fig:noise_plots}, increasing natural noise can lead to distortion in the original structure of the data in turn impacting the evaluation metrics. It can be observed that F-score (FS) decreases while Relative Distance Error (RE) increases with the increasing level of noise, which is to be expected.

\section{Conclusion} \label{sec:conclusion}

In this paper, we presented the novel privacy-preserving distance approximation framework, PPDA for distributed graph-based learning. By utilizing only the pairwise distance between clients and some set of anchors, PPDA is able to estimate the inter-client distances privately which is further used for graph learning. We also performed downstream task like graph-based clustering without accessing the actual features. Extensive experiments with both real and synthetic datasets demonstrate the efficacy of the proposed PPDA framework in preserving the structural properties of the original data. As an application to a privacy-sensitive dataset, the experiment on PANCAN produced promising results. It is, to our knowledge, the first work that can learn the graph structure for both analyzing smooth signals and building a probabilistic graphical model while respecting the privacy of the data.

\bibliographystyle{ACM-Reference-Format}
\bibliography{sample-base}

\appendix

\section{Appendix} \label{sec: appendix}

\subsection{Additional Experiment to Demonstrate PPDA} \label{sec: add_exp}
\begin{figure}[H]
     \centering
     \begin{subfigure}[b]{0.4\textwidth}
         \centering
         \includegraphics[width=\textwidth]{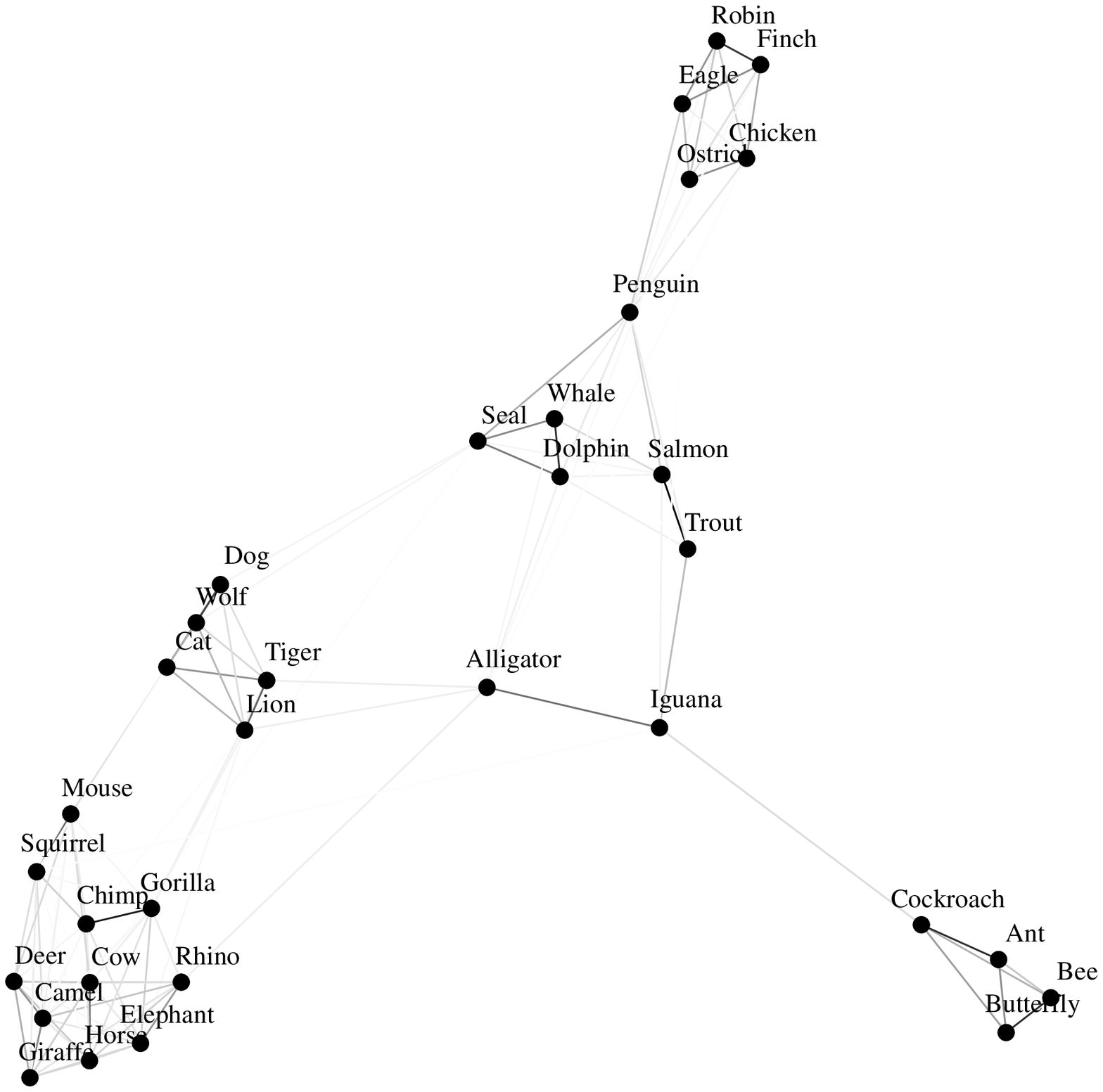}
         \caption{Non-Private}
         
     \end{subfigure}
     \hfill
     \begin{subfigure}[b]{0.4\textwidth}
         \centering
         \includegraphics[width=\textwidth]{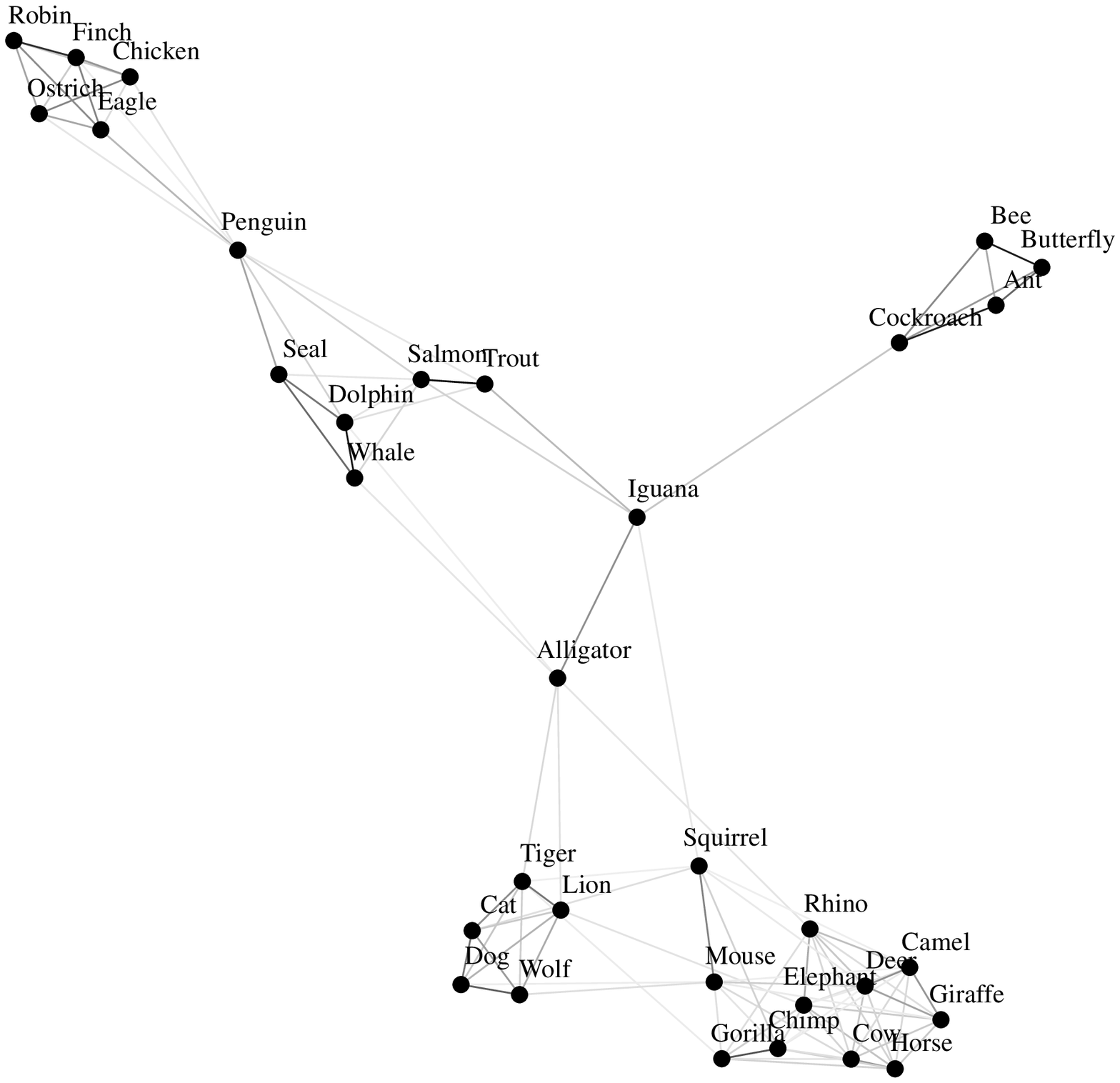}
         \caption{Private}
 
     \end{subfigure}
    
        \caption{The image depicts the graph estimated for animals dataset using SGL for (a) original features and (b) without revealing features}
        \label{fig:animals_graph}
\end{figure}

We consider the animal's dataset \cite{lake2010discovering, osherson1991default} for privacy-preserving distributed graph learning. Each animal is represented as a node in a graph and the edges between nodes indicate similarities between animals based on answers to 102 questions such as "\textit{is warm-blooded?}" and "\textit{has lungs?}". Using the proposed PPDA algorithm, we are able to preserve the neighborhood similarity with relative distance error of $0.1238$ and F-score of $0.9402$. Figure \ref{fig:animals_graph} shows the results of estimating the graph of the animals dataset using the SGL algorithm with original data and without sharing the data. The results are evaluated through visual inspection and it is expected that similar animals such as (ant, cockroach), (bee, butterfly), and (trout, salmon) are clustered together. It is clearly evident from the figure that PPDA is able to maintain the structure similarity even without sharing the data.  

\end{document}